\documentclass{article}

\PassOptionsToPackage{numbers, compress}{natbib}

\usepackage[preprint]{neurips_2025}

\usepackage[utf8]{inputenc} 
\usepackage[T1]{fontenc}    
\usepackage{hyperref}       
\usepackage{url}            
\usepackage{booktabs}       
\usepackage{amsfonts}       
\usepackage{nicefrac}       
\usepackage{microtype}      
\usepackage{xcolor}         

\usepackage{graphicx}
\usepackage{subfigure}
\usepackage{amsmath}
\usepackage{mathtools}
\usepackage{amsthm}
\usepackage{pgfplots}
\usepackage{algorithm}
\usepackage{algorithmic}
\usepackage{bm}
\usepackage{multirow}
\usepackage{makecell}
\usepackage{pifont}
\usepackage{wrapfig} 
\usepackage{caption} 

\title{SWAT: Sliding Window Adversarial Training for Gradual Domain Adaptation}

\author{%
  Zixi Wang, Xiangxu Zhao, Tonglan Xie, Mengmeng Jing, Lin Zuo \\
  School of Information and Software Engineering \\
  University of Electronic Science and Technology of China \\
}

\begin{document}

\maketitle

\begin{abstract}

Domain shifts are critical issues that harm the performance of machine learning. Unsupervised Domain Adaptation (UDA) mitigates this issue but suffers when the domain shifts are steep and drastic. Gradual Domain Adaptation (GDA) alleviates this problem in a mild way by gradually adapting from the source to the target domain using multiple intermediate domains. In this paper, we propose Sliding Window Adversarial Training (SWAT) for GDA. SWAT first formulates adversarial streams to connect the feature spaces of the source and target domains. Then, a sliding window paradigm is designed that moves along the adversarial stream to gradually narrow the small gap between adjacent intermediate domains. When the window moves to the end of the stream, i.e., the target domain, the domain shift is explicitly reduced. Extensive experiments on six GDA benchmarks demonstrate the significant effectiveness of SWAT, especially 6.1\% improvement on Rotated MNIST and 4.1\% advantage on CIFAR-100C over the previous methods.


\end{abstract}

\section{Introduction}
Traditional machine learning assumes identical training-test data distributions, yet real-world domain shifts often break this assumption and degrade model performance \cite{farahani2021brief}. Unsupervised Domain Adaptation (UDA) is proposed to mitigate domain shifts by aligning feature distributions between a labeled source domain and an unlabeled target domain \cite{pan2009survey, hoffman2018cycada}. Nevertheless, existing works \cite{Kang_2019_CVPR, tang2020discriminative, yang2020mind} have revealed that when the domain gaps are large, directly aligning two domains not only fails to reduce the domain gaps, but even causes the negative transfer \cite{pan2009survey}. 

\begin{figure}[t]
    \centering
    \includegraphics[width=0.99\columnwidth]{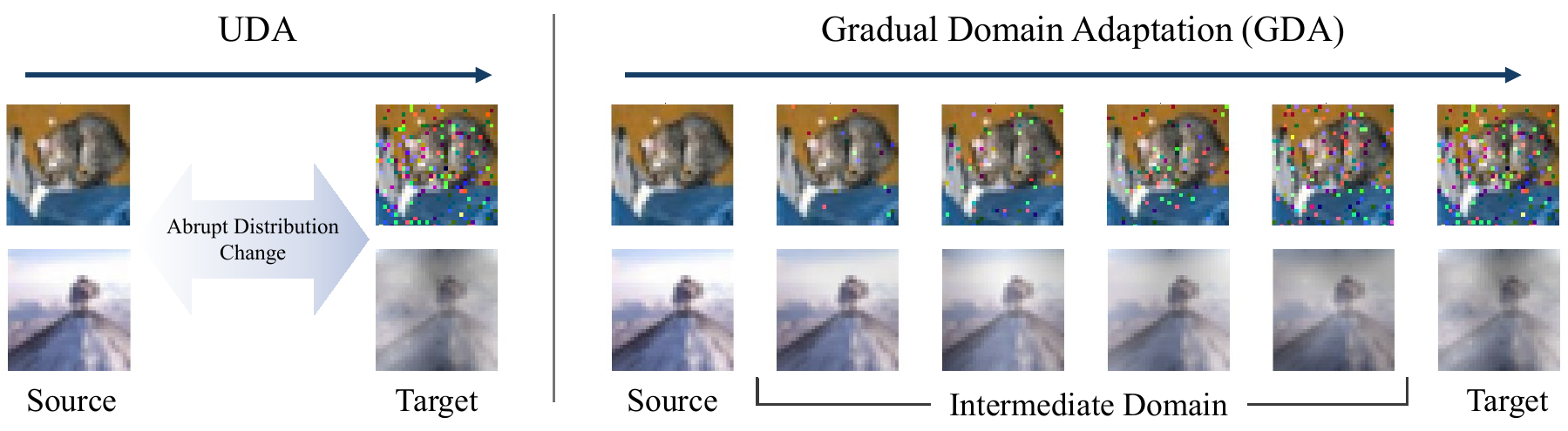}
    \caption{Comparison of UDA and GDA. Left (UDA): A single alignment maps source features directly onto the target domain. Right (GDA): Adaptation proceeds through a sequence of intermediate domains that smoothly adapt across domains, reducing abrupt distribution shifts.}
    \label{fig:gda}
\end{figure}

Gradual Domain Adaptation (GDA)\cite{kumar2020understanding} is proposed to alleviate this problem in a mild way by gradually adapting from the source to the target domain using multiple intermediate domains, as shown in Fig. \ref{fig:gda}. This paper addresses the GDA problem through adversarial training.
Adversarial training has been widely used in UDA and achieved impressive performance. This training paradigm, however, faces two challenges when applying in GDA. On the one hand, previous adversarial training methods (e.g., DANN\cite{ganin2016domain}) globally align two distributions through the game between generator and discriminator. This global matching cannot handle the continuous intermediate domains in GDA \cite{pei2018multi, shi2024adversarial}. As a result, the GDA problem degrades to the more difficult UDA problem.

On the other hand, the steep gradient of the adversarial training for large domain shifts will cause discontinuities and unsmooth problems in the manifold space \cite{rangwani2022closer, shi2024adversarial, zhang2019limitations}. As machine learning methods rely on the continuous and smooth manifold hypothesis to avoid abrupt changes in decision boundaries, this discontinuity and unsmoothness will cause error accumulation \cite{kumar2020understanding, he2023gradual, xiao2024spa}.

Towards the smooth and stable distribution matching, we propose the sliding window mechanism for adversarial learning. As a new training paradigm, the sliding window mechanism emerges three advantages over the traditional adversarial training:
(1) {\it Locality}: The sliding window mechanism avoids the globally alignment by localizing the adversarial training range, i.e., it decomposes the continuous domain flow into multiple windows, and perform adversarial training in each window to gradually refine the alignment. The generator only focuses on the distribution differences in the current window, which reduces the complexity of the adversarial training.
(2) {\it Dynamic}: The window is gradually shifted from the source domain to the target with the training process, and the update frequency of generator parameters is synchronized with the speed of domain alignment, avoiding the error accumulation caused by large optimization step in traditional adversarial training.
(3) {\it Continuity}: The sliding window continuously slides on the domain stream $H_z$, and the continuous change of parameter $p$ from 0 to 1 gradually turns the optimization focus from the left domain $H_l$ to the right domain $H_r$, avoiding the discrete switching in the multi-stage training.

Incorporated the sliding window mechanism, we present Sliding Window Adversarial Training (SWAT) method for GDA. Specifically, SWAT first formulates a bidirectional adversarial flow. This flow is optimized by a curriculum-guided sliding window, which finely controls the transition step between the source and the target domains, avoiding quadratic error accumulation caused by large transfer steps of the existing self-training strategy \cite{kumar2020understanding}. 
In the adversarial training phase, the flow generator enforces domain
continuity through sliced Wasserstein optimization across evolving domains, while the discriminator progressively filters out source-specific features through adversarial training. This co-evolutionary optimization achieves simultaneous domain invariance and target
discriminability\cite{xiao2024spa}.
The contributions are summarized as follows: 

\begin{enumerate} 
\item
We propose a sliding window mechanism to improve the adversarial training, which splits large domain shifts into multiple micro transfers through local, dynamic and continuous feature alignment, enabling fine-grained distribution matching.
\item 
We present the Sliding Window Adversarial Training (SWAT) method for GDA. SWAT can adaptively align localized domain regions, mitigating error accumulation and enabling smooth and robust knowledge transfer.
\item  
Experiments on Rotated MNIST (96.7\% vs. 90.6\% SOTA), Portraits (87.4\% vs. 86.16\% SOTA) and CIFAR-100C (24.8\% vs. 28.9\%) demonstrate the effectiveness of SWAT.
\end{enumerate}

\section{Related Work}


\textbf{Unsupervised Domain Adaptation (UDA)} aims to mitigate domain shifts by aligning feature distributions between labeled source and unlabeled target domains. Traditional approaches leverage statistical measures like Maximum Mean Discrepancy (MMD) \cite{chen2020graph} to enforce domain invariance, but face limitations under severe distribution divergence: rigid MMD-based alignment risks distorting classifier boundaries by forcibly aligning non-overlapping supports \cite{zhao2019learning}, while direct source-target alignment may erase discriminative structures, causing \textit{negative transfer} \cite{tang2020discriminative, yang2020mind}. Adversarial methods like DANN \cite{ganin2015unsupervised} and CDAN \cite{long2018conditional} advanced alignment via adversarial training but enforce fixed pairwise alignment, leading to mode collapse under disjoint supports \cite{zhao2019learning} or gradient competition under large gaps \cite{pezeshki2021gradient}. While spectral regularization \cite{pezeshki2021gradient} partially alleviates these issues, it retains rigid alignment steps. 

\textbf{Gradual Domain Adaptation (GDA)} addresses scenarios where domain shifts occur incrementally, decomposing the overall distribution gap into smaller, more manageable steps through intermediate domains \cite{farshchian2018adversarial, kumar2020understanding}. Existing methods employ diverse strategies to model these transitions: self-training leverages pseudo-labeling to bootstrap target predictions \cite{xie2020self}, gradient flow-based geodesic paths enforce smooth transitions via Riemannian manifolds \cite{zhuanggradual}, style-transfer interpolation synthesizes intermediate domains through low-level feature mixing \cite{marsden2024introducing}, and optimal transport (OT) aligns domain distributions using Wasserstein distances \cite{he2023gradual}. While alignment alone may rigidly match marginal distributions at the expense of discriminative structures. These issues are exacerbated in multi-step adaptation, where imperfectly aligned intermediates compound errors, leading to irreversible distortion of decision boundaries. Our SWAT framework uniquely preserves source-acquired information through bidirectional alignment, balancing between stability and plasticity,



\textbf{Adversarial Domain Adaptation} frameworks, including DANN \cite{ganin2015unsupervised} and CDAN \cite{long2018conditional}, have revolutionized alignment through adversarial training. These methods employ gradient reversal layers or conditional adversarial networks to learn domain-invariant representations. However, these methods enforce fixed pairwise alignment between source and target domains, leading to mode collapse when domain supports are disjoint \cite{zhao2019learning} or under large distribution gaps due to gradient competition \cite{pezeshki2021gradient}. Recent advances, such as spectral regularization \cite{pezeshki2021gradient}, partially alleviate these issues but retain the rigidity of discrete alignment steps. In contrast, our SWAT redefines domain adaptation as a \textit{continuous manifold transport process}. By constructing intermediate domains along a feature transport flow, SWAT avoids abrupt transitions and assimilates novel target modes progressively, i.e., a critical failure point for conventional UDA and adversarial methods alike.

\section{Problem Setup}

\paragraph{Domain Space}
Let $\mathcal{X} \subseteq \mathbb{R}^d$ denote the input space and $\mathcal{Y} = \{1, ..., k\}$ the label space. We model each domain as a joint probability distribution $P_t(X,Y) = P_t(X)P_t(Y|X)$ over $\mathcal{Z} = \mathcal{X} \times \mathcal{Y}$, where $t \in \{0, ..., n\}$ indexes domains along the adaptation path.

\paragraph{Gradually Shifting Domain}
In the gradually domain setting\cite{kumar2020understanding}, given a sequence of domains $\{P_t\}_{t=0}^n$ with gradually shifting distributions, where $P_0$ is the labeled source domain and $P_n$ the unlabeled target domain, GDA aims to learn a hypothesis $h: \mathcal{X} \to \mathcal{Y}$ that minimizes target risk $\epsilon_n(h)$, under two core assumptions \cite{kumar2020understanding, long2015learning}: (1) the distributional shift between consecutive domains is limited, known as Bounded Successive Divergence, and (2) the conditional distribution of labels given inputs remains unchanged across domains, referred to as Conditional Invariance.
\begin{equation}
\begin{aligned}
\text{W}_1(P_t, P_{t+1}) \leq \Delta,\quad P_t(Y|X) = P_{t+1}(Y|X), \quad \forall t \in \{0, ..., n-1\},\\
\end{aligned}
\end{equation}
where $\text{W}_1$ is the Wasserstein-1 distance and $\Delta$ quantifies maximum inter-domain drift. Conditional probability consistency ensures that label semantics remain stable during adaptation.

\paragraph{Model Pretraining in the source domain} 
The goal of pretraining in the source domain is to learn a model \( C : \mathcal{X} \to \mathcal{Y} \) that maps input features \( x \) from the training data set \( \mathcal{D} = \{(x, y)\} \) to their corresponding labels \( y \). Considering the loss function $l$, the classifier optimized on \(\mathcal{D}_t\) is denoted by \(C\), defined as:
\begin{equation}
    {C} = \arg\min_{C} \mathbb{E}_{(x, y) \sim \mathcal{D}_t} [l(C(x), y)].
\end{equation}

\paragraph{Gradual Domain Adaptation} 
Gradual domain adaptation aims to train a model $C$ that effectively generalizes to the target domain $\mathcal{D}_n$ by incrementally transferring knowledge from the labeled source domain $\mathcal{D}_0$ through a sequence of unlabeled intermediate domains $\mathcal{D}_1, \mathcal{D}_2, \dots, \mathcal{D}_{n-1}$. The adaptation process involves multi-step pseudo-labeling and self-training, where the model \( C_{0} \) is trained on the source domain and then adapted to the intermediate domains by the following self-training procedure \( \text{ST}(C_t, \mathcal{D}_t) \):
\begin{equation}
    \text{ST}(C_t, \mathcal{D}_t) = \arg\min_{C'} \mathbb{E}_{x \sim \mathcal{D}_t} [l(C'(x), \hat{y}_t(x))].
    \label{eq:st}
\end{equation}

In particular, \( \hat{y}_t(x) = \text{sign}(C_{t}(x)) \) is the pseudo-label generated by the model \( C_{t} \) for unlabeled data of \( \mathcal{D}_t \), where \( \mathcal{D}_t \) denotes the unlabeled intermediate domain. Meanwhile, \(C'\) is the next learned model, also denoted by \(C_{t+1}\).

\section{Methodology}

\begin{figure*}[t]
    \centering
    \subfigure[Feature Flow Matching]{\includegraphics[width=.49\columnwidth, trim=7cm 5cm 7cm 4.6cm, clip, page=2]{fig/feature_flow.pdf}\label{fig:feature_flow}}
    \hfill
    \subfigure[Cross-domain Adversarial Training]{\includegraphics[width=.49\columnwidth]{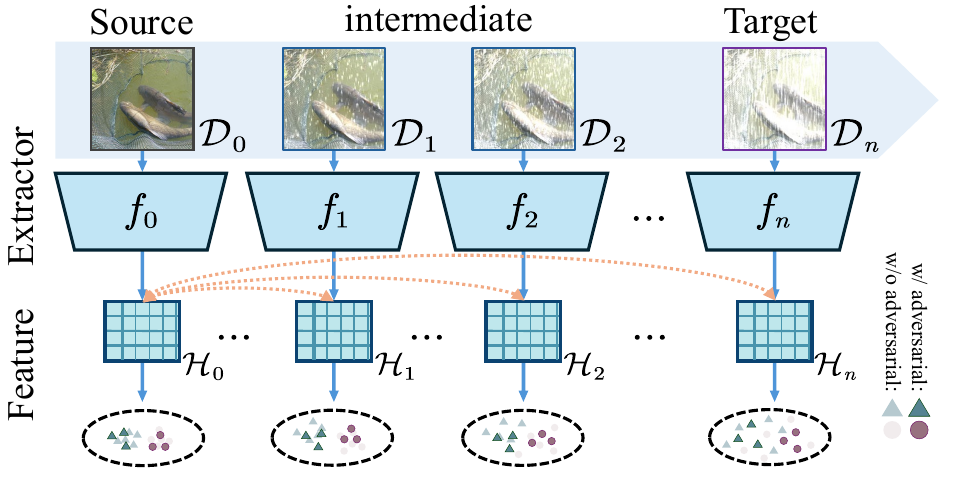}\label{fig:four-gda:subfig4}}
    \vspace{-7pt}
    \caption{(a) Illustration of sliding window mechanism, where overlapping feature spaces facilitate smooth domain transitions along the domain sequence. (b) Incremental domain alignment with the SWAT framework using adversarial training to preserve task performance and encourage feature consistency across domains. }
\end{figure*}


The proposed sliding window adversarial training models gradual domain adaptation as a continuous transition of feature distributions guided by adversarial learning. The core components are detailed in the following.

\paragraph{Continuous Feature Flow}
As illustrated in Fig.~\ref{fig:feature_flow}, SWAT first defines a continuous sequence of feature distributions:
\begin{equation}
\left\{\mathcal{H}_t\right\}_{t\in[0,n]}, \quad \mathcal{H}_t \;=\; p_t(\mathbf{h}), 
\end{equation}
over the latent space \(\mathcal H\subseteq\mathbb R^z\), where each \(\mathcal H^i\) is the feature manifold at adaptation step \(i\).  Here \(f_t:\mathcal X\to\mathcal H\) and \(g_t:\mathcal H\to\mathcal Y\) denote the encoder and classifier at step \(t\), so that the overall model \(g_t\circ f_t\) smoothly evolves from \((f_0,g_0)\) on the source domain to \((f_n,g_n)\) on the target.

Traditional GDA methods treat each pair $(f_i, g_i)$ as an independent stage in the adaptation process, where $(f_{i+1}, g_{i+1})$ is trained only after $(f_i, g_i)$ has converged. This leads to a sequence of discrete transitions between domains. In contrast, SWAT learns a continuous sequance of models \(\{(f^z,g^z)\}_{z\in[0,n]}\) and aligns features along the entire path \(\mathcal H^z\).  This continuous flow matching avoids abrupt transitions and enables fine‐grained adaptation at every intermediate point.



Unlike the discrete adaptation process in previous GDA methods, SWAT enables continuous feature transferring along the domain stream $\mathcal{H}^z$ ($z \in [0,n]$) through a sliding window, as illustrated in Fig.~\ref{fig:four-gda:subfig4}. 

\vspace{-7pt}

\paragraph{Sliding Window Mechanism}
At any step \(l\in\{0,\dots,n-1\}\), \textit{sliding window} is the pair of adjacent domains \(\{\mathcal{H}_l,\mathcal{H}_{r}\}\) with \(r=l+1\).  The scalar parameter \(p\in[0,1]\) controls where within that window we align:
\begin{equation} 
\mathcal{H}^{(l+p)} \;=\;(1-p)\,\mathcal{H}_l \;+\; p\,\mathcal{H}_{r},
\end{equation}
where \( \mathcal{H}^{(l+p)} \) refers to a domain located between \( \mathcal{H}_l \) and \( \mathcal{H}_r \). Here, $\mathcal{H}_l$ and $\mathcal{H}_r$ refer to the left and right critical domains. When \(p\) reaches 1, the window “slides” one step to the right (i.e.\ \(l \gets l+1\) and \(p\) resets toward 0), so that the next window becomes \(\{\mathcal{H}_{l+1},\mathcal{H}_{l+2}\}\).  As \(p\) varies from 0 to 1 and then triggers a slide, SWAT sweeps continuously through the entire domain stream. We then formalize the sliding-path alignment as:
\begin{equation} 
\mathcal{H}_0 \leftrightarrow \mathcal{H}^{(l + p)},\quad l \in \{0,1,\dots,n-1\},\quad p \in [0,1],
\end{equation}
where both $l$ and $p$ are parameters controlling smooth transitions across domains. This formulation enables fine-grained domain alignment through continuously shifting intermediate representations.


\paragraph{Bidirectional Flow Matching} 
Building upon the sliding window mechanism, we further incorporate it with the adversarial training for smooth flow matching. Specifically, we define \( G_m \) as the transformation function that maps a representation \( h \) from the source domain representation \( \mathcal{H}_s \) to a target domain within the domain stream \( \mathcal{H}^z \), where \( z \in [0, n] \) indicates the position of the target domain within the stream. Conversely, \( G_s \) denotes the reverse transformation, mapping features from any domain in the stream \( \mathcal{H}^z \) back to the source domain \( \mathcal{H}_s \). Thus, our SWAT model can be expressed as the bidirectional transformations:
\begin{equation} 
G_m: \mathcal{H}_s \to \mathcal{H}^z,\quad  G_s: \mathcal{H}^z \to \mathcal{H}_s .
\end{equation}

We employ the Wasserstein GAN (WGAN) \cite{2017arXiv170107875A} to train the SWAT model, as the Wasserstein distance provides a more effective measure of the distance between domains, generates higher-quality target domains \( \mathcal{H}^z \), and is easier to train. The objective function for the adversarial training module is defined as:
\begin{equation}
\min_{D} \max_{G} V(\mathbb{P}_g,\mathbb{P}_r)=\min_{D} \max_{G} \mathbb{E}_{\substack{\hat{h} \sim \mathbb{P}_g \\ h \sim \mathbb{P}_r}} \left[ D(\hat{h}) - D(h) \right] + \mathcal{R},
\label{equ:adv}
\end{equation}
where \(\hat{h}\) represents a representation generated by the generator \( G \), which approximates the target domain distribution \(\mathbb{P}_g\). \(h\) is a representation from the real data distribution \(\mathbb{P}_r\), corresponding to actual data from the target domain. \(D\) denotes the discriminator of the corresponding domain, and different domains have different discriminators. \(\mathcal{R}\) represents the regularization term proposed by \citet{2017arXiv170400028G}:
\begin{equation}
\mathcal{R} = \mathbb{E}_{\tilde{h} \sim \mathbb{P}_{\tilde{h}}} \left[ \lambda \left( || \nabla_{\tilde{h}} D(\tilde{h}) ||_2 - 1 \right)^2 \right],
\end{equation}
where \( \tilde{h} \) denotes a random linear interpolation of points from $\hat{h}$ and $h$ representations, and \( \lambda \) is a hyperparameter controlling the strength of the regularization.

To facilitate bidirectional feature alignment between the source domain $\mathcal{H}_0$ and the critical domains, we formulate bidirectional flow matching based on the minimax objective $V(\mathbb{P}_g,\mathbb{P}_r)$ defined in Eq. (\ref{equ:adv}). Without loss of generality, taking the left critical domain $H_l$ as an example, the adversarial loss enforces cross-domain distribution matching through dual mapping paths:
\begin{equation}
\mathcal{L}^l_{\text{adv}} = V\left(G_m(\mathcal{H}_0),\mathcal{H}_l\right) + V\left(G_s(\mathcal{H}_l),\mathcal{H}_0\right),
\end{equation}
\label{eq:adv}
where $G_m$ maps source features to the critical domain while $G_s$ reconstructs the original domain. The symmetrical adversarial loss $\mathcal{L}^r_{\text{adv}}$ for the right critical domain $\mathcal{H}_r$ follows the same dual-path formulation.

\paragraph{Semantic Consistency Preservation}
To prevent mode collapse and maintain content integrity during adaptation, we employ cycle-consistent constraints inspired by CycleGAN. This ensures that features cyclically transformed through $\mathcal{H}_0 \rightarrow \mathcal{H}_l \rightarrow \mathcal{H}_0$ to preserve semantic consistency:
\begin{equation}
\begin{gathered}
\mathcal{L}^l_{\text{cycle}} = 
   \mathbb{E}_{h\sim\mathcal{H}_{0}}\left[\|G_s(G_m(h)) - h\|_2\right]
   +\mathbb{E}_{h\sim\mathcal{H}_{l}}\left[\|G_m(G_s(h)) - h\|_2\right].
\end{gathered}
\label{eq:cycle}
\end{equation}

The bidirectional reconstruction regularizations enforce invertible transformations while penalizing semantic distortions, particularly crucial for preserving task-relevant features in critical domains.

\subsection{The Overall Objective}


Similar to previous GDA methods, we also optimize the self-training loss as follows:
\begin{equation}
    \mathcal{L}^l_{st} = \mathbb{E}_{h \sim \mathcal{H}} [l(g(h), \hat{y}(h))],
\end{equation}

\begin{wrapfigure}{r}[0cm]{0pt}        
    \includegraphics[width=.36\columnwidth, trim=7.5cm 6.1cm 7.5cm 6cm, clip, page=1]{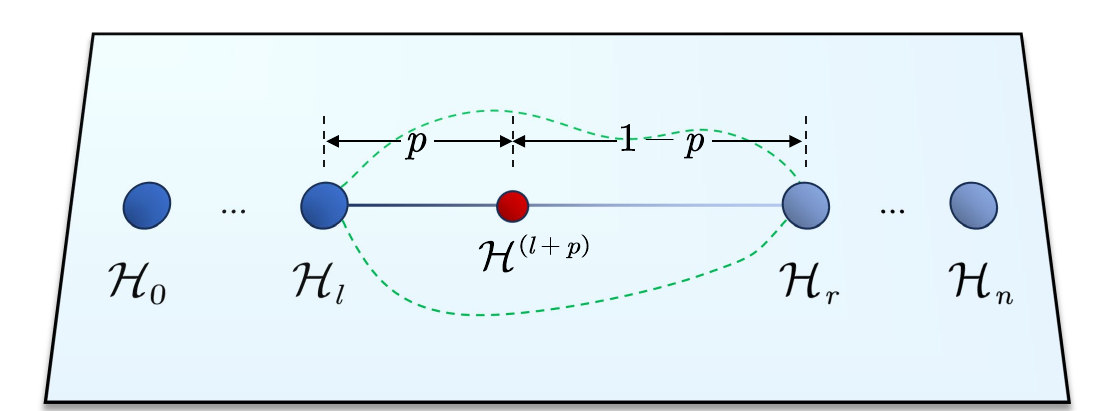}
    \caption{\(\mathcal{H}^{(l+p)}\) lies along the shortest domain flow determined by \(p\).}
    \label{fig:dist}
\end{wrapfigure}

where $l$ is the cross-entropy loss. When \( h \) comes from the unlabeled domain, \( \hat{y}_t(x) \) is the pseudo-label generated by the model \( g \). But when \( h \) is a feature generated by \( G_m(h_0) \), it represents the ground-truth label of the original representation \( h_0 \) from the source domain.

As shown in Fig. \ref{fig:dist}, the generated feature space $\mathcal{H}^{(l+p)}$ is enforced to satisfy the condition $\text{dist}(\mathcal{H}_l, \mathcal{H}^{(l+p)})$/$\text{dist}(\mathcal{H}_r, \mathcal{H}^{(l+p)})$$=$$p/(1-p)$, where $\text{dist}(\cdot, \cdot)$ denotes a valid distance metric between two distributions. 

The overall objective of SWAT is formulated as follows:
\begin{equation}
\mathcal{L} = (1-p)\mathcal{L}^l+p\mathcal{L}^r,
\label{eq:total}
\end{equation}

where $\mathcal{L}_l$ is the adversarial training loss which defines as: $\mathcal{L}^l = \mathcal{L}^l_{adv} + \mathcal{L}^l_{cycle} + \mathcal{L}^l_{st}$. By optimizing Eq. \ref{eq:total}, we achieve continuous flow matching in the feature space. For clear understanding, we summarize the main idea of SWAT in Algorithm \ref{alg:flow-gta}. 

\begin{algorithm}[h]
\caption{Sliding Window Adversarial Training (SWAT)}
\label{alg:flow-gta}
\begin{algorithmic}[1]
\STATE {\bfseries Input:} A series of domains $\mathcal{D}_0\text{(source)},$ $\mathcal{D}_1,$ $\ldots,$ $\mathcal{D}_{n-1}$ $\mathcal{D}_n\text{(target)}$, pretrained encoder $f$ and classifier $g$.
\STATE {\bfseries Output:} $\frac{1}{N} \sum_{i=1}^{N} 1(\hat{y}_i = \mathcal{Y}_i)$, where $\hat{y} = g(f(\mathcal{D}_n))$ 
\STATE Initialize generators $G_m, G_s$ and discriminators $D_s, D_l, D_r$.
\FOR{$l = 0$ \textbf{to} $n-1$}
    \STATE $r \gets l + 1$
    \STATE $D_l \gets D_r$, and reinitialize $D_r$.
    \STATE $\mathcal{H}_0 \gets f(\mathcal{D}_0), \mathcal{H}_l \gets f(\mathcal{D}_l), \mathcal{H}_r \gets f(\mathcal{D}_r)$
    \FOR{$p = 0$ \textbf{to} $1$, with step size $\Delta p$}  
        \STATE $\mathcal{L} = (1-p)\cdot L(G_m, G_s, D_s, D_l, \mathcal{H}_0, \mathcal{H}_l) + p \cdot L(G_m, G_s, D_s, D_r, \mathcal{H}_0, \mathcal{H}_r)$ 
        \STATE update $G_m, G_s, D_s, D_l, D_r, f, g$
    \ENDFOR
\ENDFOR
\end{algorithmic}
\end{algorithm}

\section{Experiments}

\subsection{Datasets and Implementation Details}

Following the standard GDA protocol, we conduct extensive experiments on 6 datasets. 
\textbf{Rotated MNIST} is constructed from MNIST \cite{deng2012mnist}, this dataset contains 50,000 source domain images (original digits) and 50,000 target domain images rotated by 45°. Intermediate domains interpolate rotation angles between 0° and 45°. 
\textbf{Color-Shift MNIST} images are normalized to [0,1] for the source domain and shifted to [1,2] for the target domain \cite{he2023gradual}, with intermediate domains generated by linearly interpolating color intensity. 
\textbf{Portraits \cite{ginosar2015century}} are chronologically divided into 9 temporal domains (1905–2013), each with 2,000 images \cite{kumar2020understanding}. The first and last domains serve as source/target; images are resized to 32×32 pixels.
\textbf{Cover Type\cite{covertype_31}} tabular dataset sorted by horizontal distance to water, uses 50,000 source samples, 10×40,000 intermediate domains, and 50,000 target samples \cite{kumar2020understanding} for classifying spruce fir vs. Rocky Mountain pine.

To evaluate the performance of GDA methods under high-severity shifts, we introduce a new evaluation protocol using the corruption benchmarks \textbf{CIFAR-10C}, \textbf{CIFAR-100C}~\cite{hendrycks2019benchmarking}. Each benchmark applies 15 corruption types at five severity levels to the validation and test splits of CIFAR~\cite{krizhevsky2009learning}. We regard the clean training images as the source domain, the images of severity levels 1–4 (across all corruption types) as a sequence of intermediate domains, and treat severity level 5 as the target domain. Following the RobustBench benchmark~\cite{croce2020robustbench, croce2020reliable}, WideResNet-28~\cite{zagoruyko2016wide} and ResNeXt-29~\cite{xie2017aggregated} used as the source model for CIFAR10-to-CIFAR10C and CIFAR100-to-CIFAR100C, respectively.

All results are averaged over 5 runs. Please refer to Appendix \ref{sec:appendix:result} for more detailed implementation.

\subsection{Experimental Results}
\label{sec:results}

\begin{table*}[h]
\newcommand{\mini}[1]{\footnotesize{(#1)}}
\centering
\caption{Comparison of domain adaptation methods on 4 GDA datasets.}
\renewcommand{\arraystretch}{1.0} 
\resizebox{0.99\columnwidth}{!}{%
\begin{tabular}{lccccc}
\toprule Methods & Gradual & {Rotated MNIST} & {Color-Shift MNIST} & Portraits & Cover Type \\
\midrule
{DANN \cite{ganin2016domainneural}} & \ding{55} & $44.2$ & $56.5$ & $73.8$ & - \\
{DeepCoral \cite{sun2016deep}} & \ding{55} & $49.6$ & $63.5$ & $71.9$ & - \\
{DeepJDOT \cite{damodaran2018deepjdot}} & \ding{55} & $51.6$ & $65.8$ & $72.5$ & - \\
\midrule
GST \cite{kumar2020understanding} \mini{ICML'20} & \ding{51} & $83.8$ & $74.0$ & $82.6$ & $73.5$ \\
IDOL \cite{chen2021gradual} \mini{NeurIPS'21} & \ding{51} & $87.5$ & - & $85.5$ & - \\
AGST \cite{zhou2022active} \mini{IEEE'22} & \ding{51} & 76.2 & - & 77.6 &\\
GGF \cite{zhuanggradual} \mini{ICLR'24} & \ding{51} & $67.7$ & - & $86.2$ & -\\
GOAT \cite{he2023gradual} \mini{JMLR'24} & \ding{51} & $86.4$ & $91.8$ & $83.6$ & $69.9$ \\
DRO \cite{najafi2024gradual} \mini{NeurIPS'24} & \ding{51} & $53.2$ & - & - & - \\
AST \cite{shi2024adversarial} \mini{NeurIPS'24} & \ding{51} & $90.6$ & - & $84.8$ & - \\
CNF \cite{sagawa2025gradual} \mini{Neural Computation'25} & \ding{51} & $62.6$ & - & $84.6$ & - \\
SWAT \mini{Ours} & \ding{51} & $\mathbf{96.7}$ & $\mathbf{99.6}$ & $\mathbf{87.4}$ & $\mathbf{75.0}$ \\
\bottomrule
\end{tabular}%
}
\label{tab:comp-uda}
\end{table*}

Compared with the UDA methods \cite{he2023gradual}, table \ref{tab:comp-uda} highlights the clear advantage of gradual domain adaptation (GDA) over traditional unsupervised domain adaptation (UDA) and reveals the clear performance advantages in the GDA setting: Whereas UDA methods like DANN, DeepCORAL, and DeepJDOT struggle under large domain shifts, achieving at best 51.6\% on Rotated MNIST and 65.8\% on Color-Shift MNIST, these results highlight SWAT’s ability to learn smooth and discriminative representations under gradual shifts.

\begin{table*}[h]
  \centering
  \caption{Comparison of SWAT and other GDA methods on 4 datasets given different numbers of intermediate domains.}
  \setlength{\tabcolsep}{5pt} 
  \renewcommand{\arraystretch}{1.1} 
  \resizebox{0.9\textwidth}{!}{
  \begin{tabular}{c|ccc|c|ccc}
    \hline
    \multirow{2}{*}{\rotatebox{0}{\makecell{Given \\Domains}}} & \multicolumn{3}{c|}{Rotated MNIST} & \multirow{2}{*}{\rotatebox{0}{\makecell{Given \\Domains}}} & \multicolumn{3}{c}{Color-Shift MNIST}\\
    & GST & GOAT & SWAT & & GST & GOAT & SWAT \\
    \hline
    2 & $54.9 \pm 0.2$ & $53.5 \pm 1.0$ & $\mathbf{88.1 \pm 1.5}$ &
    2 & $27.0 \pm 0.3$ & $72.0 \pm 6.0$ & $\mathbf{98.8 \pm 0.3}$ \\
    3 & $60.0 \pm 0.3$ & $57.2 \pm 0.3$ & $\mathbf{96.1 \pm 0.1}$ &
    3 & $34.2 \pm 1.7$ & $83.4 \pm 2.9$ & $\mathbf{99.5 \pm 0.0}$ \\
    4 & $67.2 \pm 0.6$ & $68.4 \pm 1.4$ & $\mathbf{96.4 \pm 0.0}$ &
    4 & $55.0 \pm 1.9$ & $89.1 \pm 3.6$ & $\mathbf{99.6 \pm 0.0}$ \\
    5 & $71.9 \pm 0.8$ & $78.8 \pm 0.8$ & $\mathbf{96.5 \pm 0.2}$ &
    5 & $66.8 \pm 2.2$ & $94.9 \pm 1.0$ & $\mathbf{99.6 \pm 0.0}$ \\
    6 & $75.6 \pm 1.4$ & $85.8 \pm 0.9$ & $\mathbf{96.7 \pm 0.1}$ &
    6 & $74.0 \pm 3.4$ & $95.7 \pm 0.3$ & $\mathbf{99.6 \pm 0.0}$ \\
    \hline
    \multicolumn{3}{c}{}\\ 
    \hline
    \multirow{2}{*}{\rotatebox{0}{\makecell{Given \\Domains}}} & \multicolumn{3}{c|}{Portraits} & \multirow{2}{*}{\rotatebox{0}{\makecell{Given \\Domains}}}  & \multicolumn{3}{c}{Cover Type} \\
    & GST & GOAT & SWAT & & GST & GOAT & SWAT  \\
    \hline
    2 & $75.0 \pm 1.7$ & $78.6 \pm 2.2$ & $\mathbf{85.3 \pm 0.1}$ &
    2 & $69.1 \pm 0.1$ & $69.0 \pm 0.0$ & $\mathbf{75.0 \pm 0.0}$ \\
    3 & $75.1 \pm 1.0$ & $80.2 \pm 1.3$ & $\mathbf{84.8 \pm 1.0}$ &
    3 & $71.1 \pm 0.2$ & $69.0 \pm 0.0$ & $\mathbf{74.3 \pm 0.2}$ \\
    4 & $78.4 \pm 0.9$ & $80.5 \pm 1.3$ & $\mathbf{86.1 \pm 0.3}$ &
    4 & $72.4 \pm 0.1$ & $69.0 \pm 0.0$ & $\mathbf{74.6 \pm 0.1}$ \\
    5 & $76.4 \pm 1.8$ & $79.4 \pm 0.6$ & $\mathbf{87.0 \pm 0.0}$ &
    5 & $72.8 \pm 0.1$ & $69.1 \pm 0.1$ & $\mathbf{74.6 \pm 0.1}$ \\
    6 & $80.9 \pm 0.6$ & $83.1 \pm 0.6$ & $\mathbf{87.4 \pm 0.2}$ &
    6 & $\mathbf{73.1 \pm 0.1}$ & $69.3 \pm 0.0$ & $\mathbf{73.7 \pm 0.2}$\\
    \hline
  \end{tabular}
  }
  \label{tab:comp1}
\end{table*}

Table \ref{tab:comp1} compares SWAT against GST \cite{kumar2020understanding} and GOAT \cite{he2023gradual} on both vision benchmarks (Rotated MNIST, Color-Shift MNIST, Portraits) and the Cover Type tabular dataset, using the same encoder–classifier architecture and low-confidence sample selection strategy throughout. SWAT consistently outperforms GST and GOAT across every setting, with the largest gains observed when only two or three domains are available. Narrow confidence intervals further confirm the stability of our results. By more effectively leveraging domain flow and feature transfer SWAT delivers superior adaptation across diverse data modalities. (Additional experiments and computational-cost comparisons are provided in Appendices \ref{sec:appendix:Implementation} and \ref{sec:appendix:cost}, respectively.)

\begin{table*}[h]
\centering
\caption{Comparison of classification error rates (\%) at severity level 5 for test‐time adaptation (TTA) and gradual domain adaptation (GDA) methods on CIFAR-10C and CIFAR-100C. Lower is better.}
\setlength{\tabcolsep}{3pt} 
\renewcommand{\arraystretch}{1.1} 
\resizebox{\textwidth}{!}{
\begin{tabular}{c|l|c|ccccccccccccccc|c}
\hline
\rotatebox{90}{} & \raisebox{3.5ex}{Method} & \rotatebox{90}{Gradual} & \rotatebox{70}{\makecell{gaussian}} & \rotatebox{70}{\makecell{shot}} & \rotatebox{70}{impulse} & \rotatebox{70}{defocus} & \rotatebox{70}{glass} & \rotatebox{70}{motion} & \rotatebox{70}{zoom} & \rotatebox{70}{snow} & \rotatebox{70}{frost} & \rotatebox{70}{fog} & \rotatebox{70}{brightness} & \rotatebox{70}{contrast} & \rotatebox{70}{elastic} & \rotatebox{70}{pixelate} & \rotatebox{70}{jpeg} & \raisebox{3.5ex}{Mean} \\ 
\hline
\multirow{9}{*}{\rotatebox{90}{CIFAR-10C}}
&Source only&\ding{55}&72.3&65.7&72.9&46.9&54.3&34.8&42.0&25.1&41.3&26.0&9.3&46.7&26.6&58.5&30.3&43.5\\
&BN-1       &\ding{55}&28.1&26.1&36.3&12.8&35.3&14.2&12.1&17.3&17.4&15.3&8.4&12.6&23.8&19.7&27.3&20.4\\
&TENT-cont. \cite{wang2020tent}&\ding{55}&24.8&20.6&28.6&14.4&31.1&16.5&14.1&19.1&18.6&18.6&12.2&20.3&25.7&20.8&24.9&20.7\\
&AdaContrast \cite{chen2022contrastive}&\ding{55}&29.1&22.5&30.0&14.0&32.7&14.1&12.0&16.6&14.9&14.4&8.1&10.0&21.9&17.7&20.0&18.5\\
&CoTTA \cite{wang2022continual}&\ding{55}&24.3&21.3&26.6&11.6&27.6&12.2&10.3&14.8&14.1&12.4&7.5&10.6&\textbf{18.3}&13.4&\textbf{17.3}&16.2\\
&GTTA-MIX \cite{marsden2022gradual}&\ding{55}&23.4&\textbf{18.3}&\textbf{25.5}&10.1&\textbf{27.3}&11.6&10.1&14.1&\textbf{13.0}&\textbf{10.9}&7.4&9.0&19.4&14.5&19.8&15.6\\
&GST \cite{kumar2020understanding}&\ding{51}&50.0&43.9&50.3&20.6&51.2&17.2&16.7&17.5&24.3&17.5&6.9&13.2&24.9&39.9&26.6&28.1\\
&GOAT \cite{he2023gradual}&\ding{51}&72.7&65.7&73.0&46.7&54.5&34.3&41.5&24.9&41.0&26.0&9.3&46.6&26.4&58.1&30.2&43.4\\
&SWAT(ours)  &\ding{51}&\textbf{21.4}&20.0&26.8&\textbf{9.7}&28.5&\textbf{10.2}&\textbf{8.4}&\textbf{3.1}&13.4&11.5&\textbf{7.1}&\textbf{8.9}&19.8&\textbf{13.3}&20.1&\textbf{15.4}\\
\hline
\multirow{9}{*}{\rotatebox{90}{CIFAR-100C}}
&Source only&\ding{55}&73.0&68.0&39.4&29.3&54.1&30.8&28.8&39.5&45.8&50.3&29.5&55.1&37.2&74.7&41.2&46.4\\
&BN-1       &\ding{55}&42.1&40.7&42.7&27.6&41.9&29.7&27.9&34.9&35.0&41.5&26.5&30.3&35.7&32.9&41.2&35.4\\
&TENT-cont. \cite{wang2020tent}&\ding{55}&37.2&35.8&41.7&37.9&51.2&48.3&48.5&58.4&63.7&71.1&70.4&82.3&88.0&88.5&90.4&60.9\\
&AdaContrast \cite{chen2022contrastive}&\ding{55}&42.3&36.8&38.6&27.7&40.1&29.1&27.5&32.9&30.7&38.2&25.9&28.3&33.9&33.3&36.2&33.4\\
&CoTTA \cite{wang2022continual}&\ding{55}&40.1&37.7&39.7&26.9&38.0&27.9&26.4&32.8&31.8&40.3&24.7&26.9&32.5&28.3&33.5&32.5\\
&GTTA-MIX \cite{marsden2022gradual}&\ding{55}&36.4&32.1&34.0&24.4&35.2&25.9&23.9&28.9&27.5&30.9&22.6&23.4&29.4&25.5&33.3&28.9\\
&GST \cite{kumar2020understanding}&\ding{51}&49.8&56.7&32.3&22.5&41.6&25.0&23.3&30.3&32.2&38.1&22.1&27.0&33.1&40.8&35.8&33.3\\
&GOAT \cite{he2023gradual}&\ding{51}&73.4&67.9&39.1&28.7&53.8&30.2&28.7&39.3&45.7&50.0&29.4&53.7&36.8&74.3&41.2&46.2\\
&SWAT(ours)  &\ding{51}&\textbf{28.6}&\textbf{26.9}&\textbf{23.5}&\textbf{22.3}&\textbf{29.0}&\textbf{22.7}&\textbf{22.4}&\textbf{24.4}&\textbf{24.3}&\textbf{25.7}&\textbf{21.5}&\textbf{22.7}&\textbf{26.5}&\textbf{23.4}&\textbf{28.8}&\textbf{24.8}\\
\hline
\end{tabular}
}
\label{tab:result-comp}
\end{table*}

Table \ref{tab:result-comp} presents classification error rates (severity level 5) on CIFAR-10C and CIFAR-100C. We group methods into two families: Test-Time Adaptation (TTA) and gradual domain adaptation (GDA). For TTA, “Source only” refers to the fixed pretrained model, and BN-1 updates batch normalization statistics on each test batch. The other baselines, including TENT-continual \cite{wang2020tent}, AdaContrast \cite{chen2022contrastive}, CoTTA \cite{wang2022continual}, and GTTA-MIX \cite{marsden2022gradual}, perform online adaptation of either the feature extractor or the classifier. Our approach, SWAT, combines the stability of batch-norm re-estimation with sample-wise alignment. Across both benchmarks, SWAT attains the lowest mean error (15.4\% on CIFAR-10C, 24.8\% on CIFAR-100C), outperforming the strongest prior TTA (GTTA-MIX: 15.6\%/28.9\%) and GDA competitors on nearly every corruption type. 


\subsection{Domain Shifts Analysis} 

\begin{figure}[h]
\centering
\definecolor{mycolor1}{HTML}{CC5F5A}
\definecolor{mycolor2}{HTML}{E6C786}
\definecolor{mycolor3}{HTML}{82ABA3}
\begin{tikzpicture}
\begin{axis}[
    width=3.5cm,
    height = 3.5cm,
    ylabel={$\mathcal{A}$-distance},
    ylabel style={yshift=-10pt},
    grid=major,
]
\addplot[
    color=mycolor1, 
    mark=*,
    thick
] 
coordinates {
    (0, 0) (1/3, 1.078) (2/3, 1.18) (1, 1.284)
};
\addplot[
    color=mycolor2, 
    mark=*,
    thick
] 
coordinates {
    (0, 1.284) (1/3, 1.498) (2/3, 1.12) (1, 0.0)
};

\end{axis}
\end{tikzpicture}
\begin{tikzpicture}
\begin{axis}[
    width=7cm,
    height=3.5cm,
    grid=major,
    legend style={at={(1.05,0)}, anchor=south west, legend columns=1, column sep=1ex, row sep=1ex, draw=none},
    scaled y ticks=false,          
    y tick label style={
        /pgf/number format/fixed,  
        /pgf/number format/precision=2  
    },
]
\addplot[
    color=mycolor1,
    mark=*,
    thick
] 
coordinates {
    (0, 0.0) (1/9, 0.090) (2/9, 0.090) (3/9, 0.082) (4/9, 0.094) (5/9, 0.094) (6/9, 0.12) (7/9, 0.112) (8/9, 0.112) (1, 0.104)
};
\addlegendentry{distance to $\mathcal{H}_0$}

\addplot[
    color=mycolor2, 
    mark=*,
    thick
] 
coordinates {
    (0, 0.104) (1/9, 0.064) (2/9, 0.062) (3/9, 0.094) (4/9, 0.102) (5/9, 0.052) (6/9, 0.110) (7/9, 0.070) (8/9, 0.088) (1, 0.0)
};
\addlegendentry{distance to $\mathcal{H}_n$}

\addplot[
    color=mycolor3, 
    mark=*,
    thick
] 
coordinates {
    (0, 0.0) (1/9, 0.0) (2/9, 0.0) (3/9, 0.0) (4/9, 0.0) (5/9, 0.0) (6/9, 0.0) (7/9, 0.0) (8/9, 0.0) (1, 0.0)
};
\addlegendentry{distance to $\mathcal{H}_l,\mathcal{H}_r$}
\end{axis}
\end{tikzpicture}
\vspace{-7pt}
\caption{The figure illustrates how the interpolated domains evolve in domain discrepancy along the adaptation path. The horizontal axis denotes the interpolation place~$z\in[0,1]$ (0 = source domain, 1 = target domain). The vertical axis represents the $\mathcal{A}$-distance \cite{ben2010theory,mansour2009domain}, a proxy for distribution divergence. On the left, the $\mathcal{A}$-distance is computed with representations of a fixed encoder, while on the right, the distance is calculated using our SWAT representation.}
\label{fig:a-distance}
\end{figure}

\paragraph{Quantitative Analysis of Domain Shifts}
We employ $\mathcal{A}$-distance \cite{ben2010theory} as the proxy of $\mathcal{H} \Delta \mathcal{H}$ distance to quantitatively evaluate the domain shifts, as illustrated in Fig. \ref{fig:a-distance}. We observe that the $\mathcal{A}$-distance between the source domain $H_0$ and the target domain $H_n$ exhibits large fluctuations (peak at 1.498), which indicates that directly aligning two domains causes unstable transfer or even negative transfer when the domain shifts are significantly large.
In contrast, the proposed SWAT maintains near-zero distances ($< 0.11$) to critical intermediate domains $\mathcal{H}_l, \mathcal{H}_r$ across all positions, achieving a 63.7\% reduction in the average $\mathcal{A}$-distance between $\mathcal{H}_0$ and $\mathcal{H}_n$ (0.104 vs. 1.284), demonstrating smooth knowledge transfer. The symmetrical reduction of bidirectional distances confirms balanced adaptation between forward and backward domain transitions.

\begin{figure*}[h]
    \centering
    \subfigure[Source space $\mathcal{H}_0$]{\includegraphics[width=.19\textwidth]{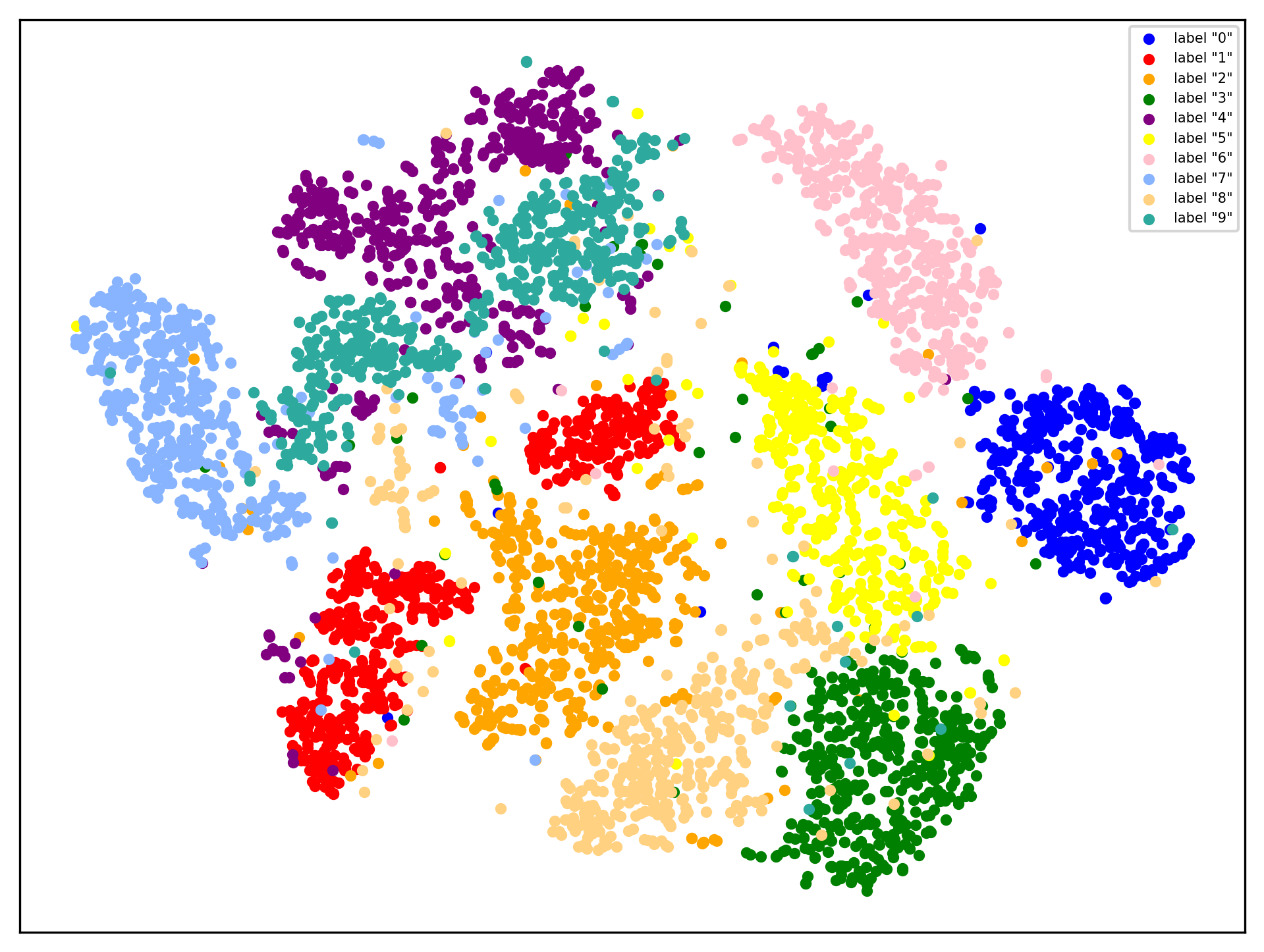}\label{fig:feature_encoder:a}}
    \subfigure[Target space $\mathcal{H}_n$ w/o flow matching]{\includegraphics[width=.19\textwidth]{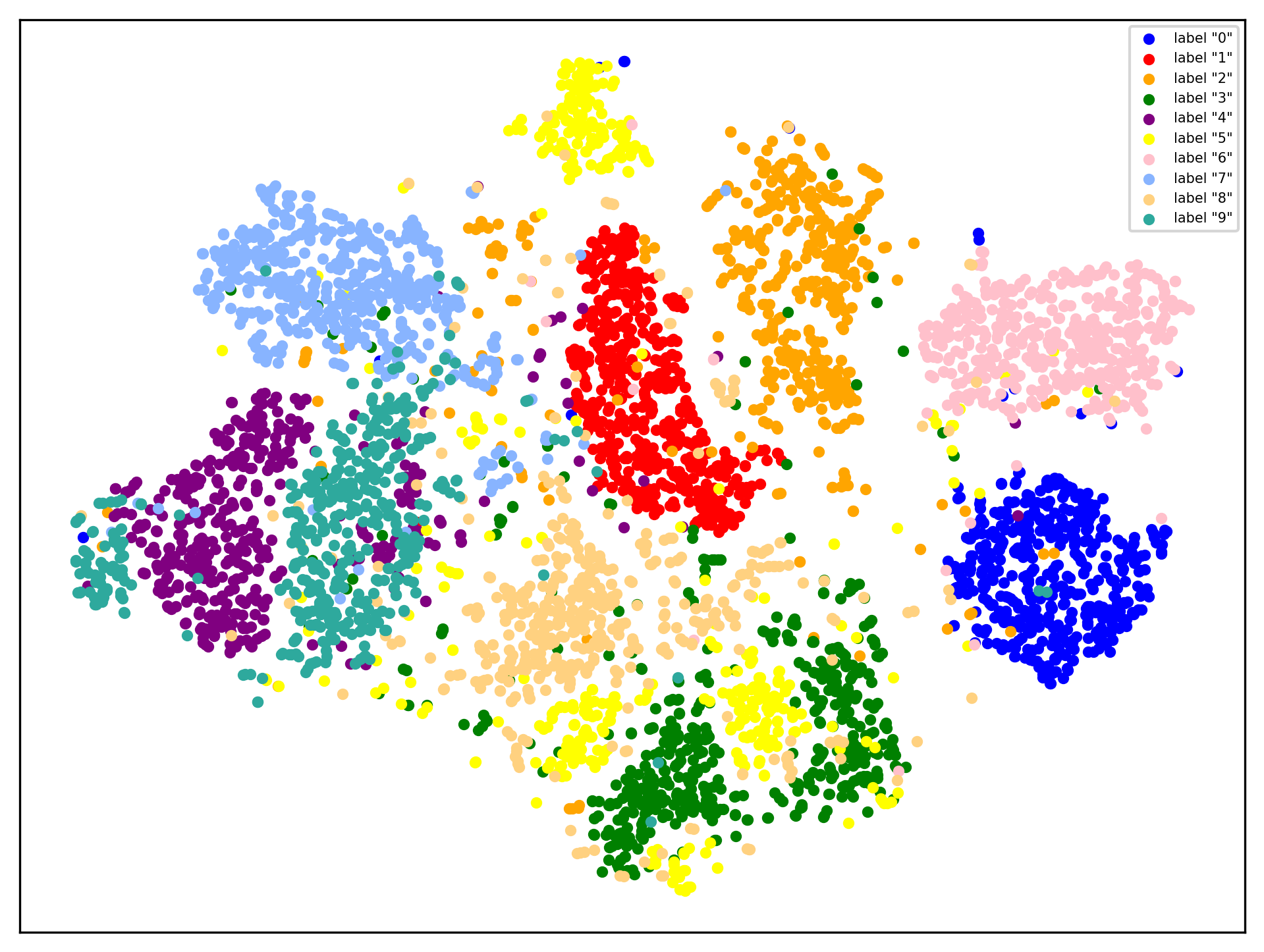}\label{fig:feature_encoder:b}}
    \subfigure[Target space $\mathcal{H}_n$ w/ flow matching]{\includegraphics[width=.19\textwidth]{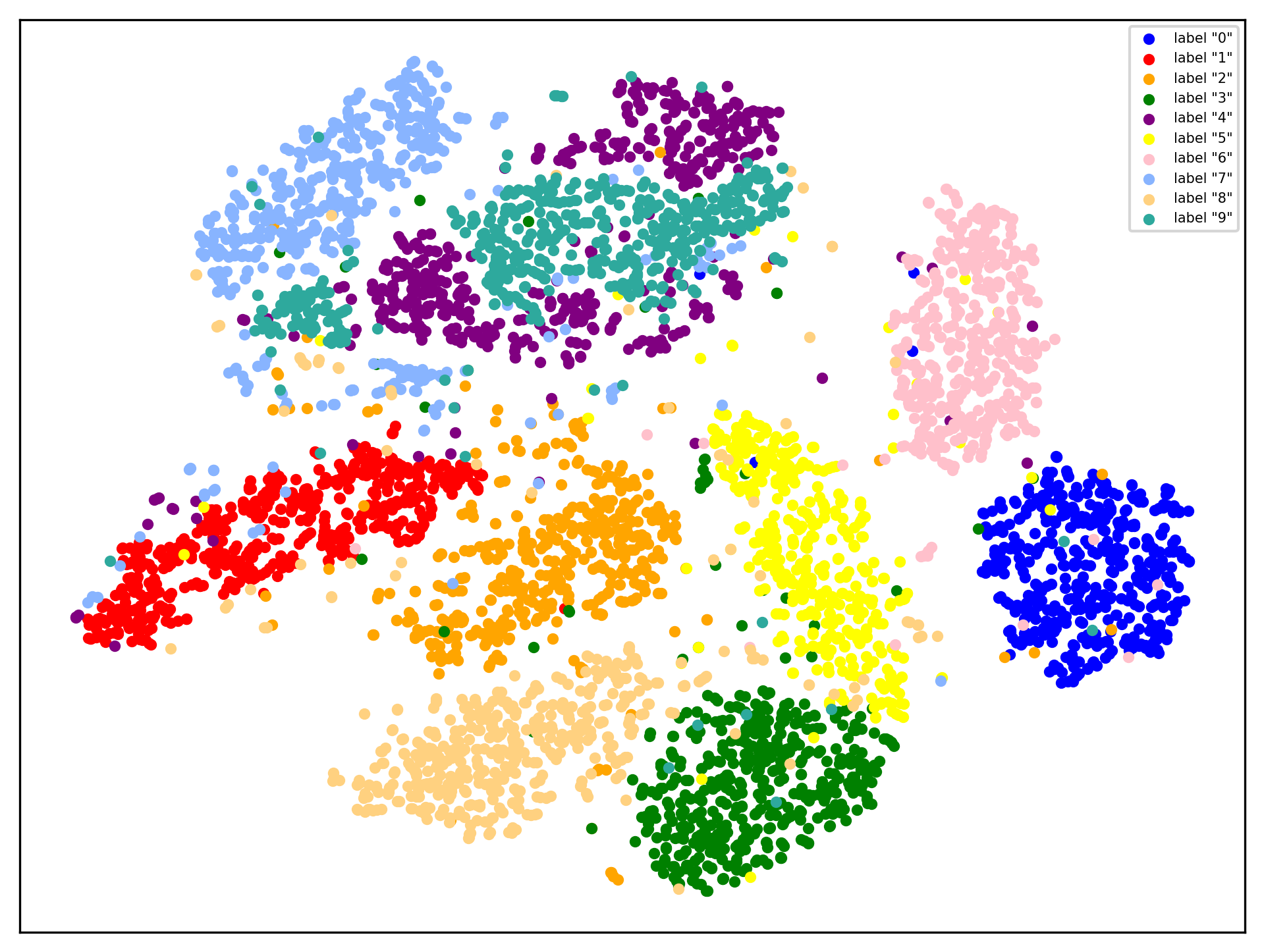}\label{fig:feature_encoder:c}}
    \subfigure[$\mathcal{H}_0$,$\mathcal{H}_n$ w/o adversarial training]{\includegraphics[width=.19\textwidth]{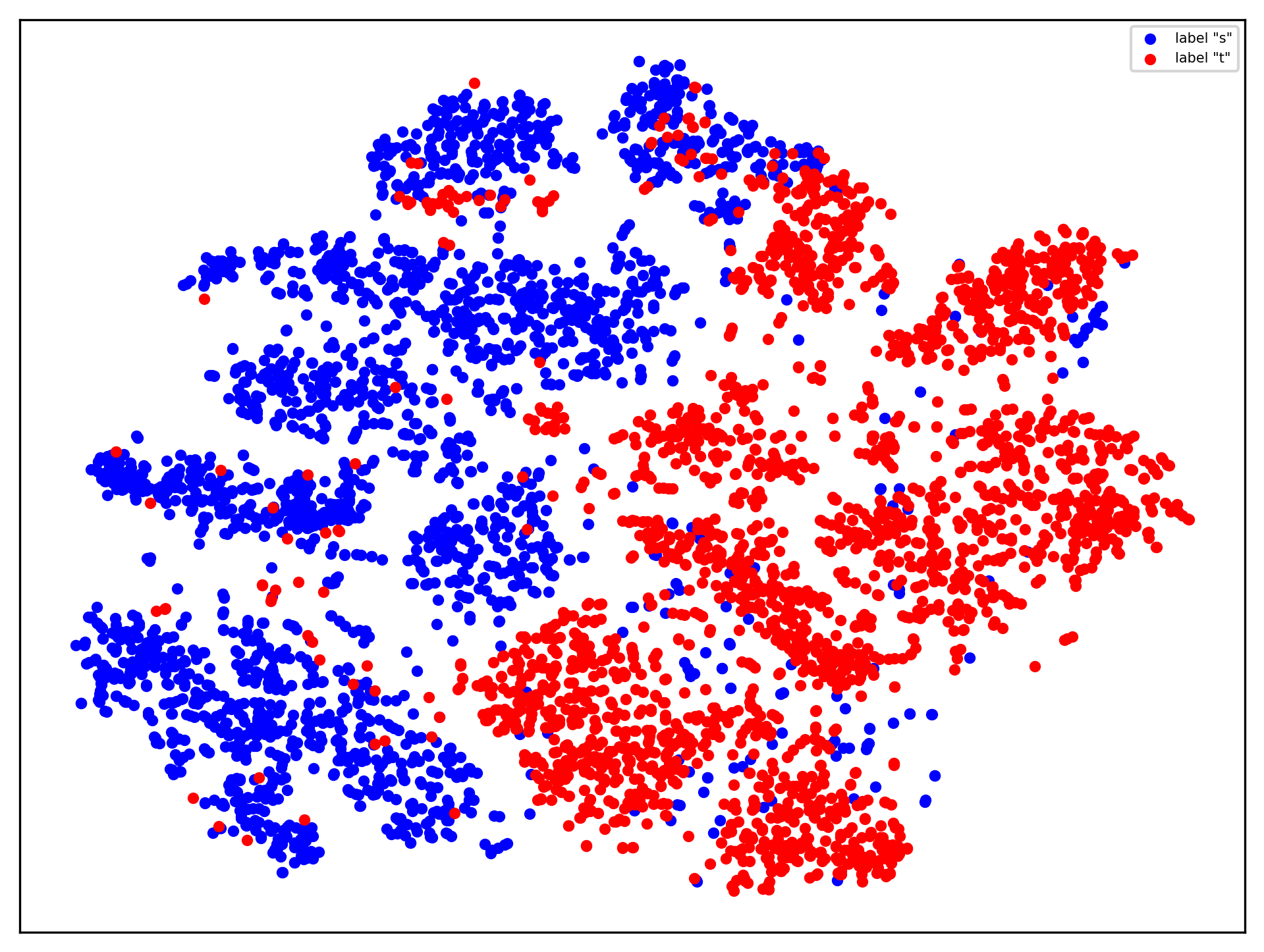}\label{fig:feature_encoder:d}}
    \subfigure[$\mathcal{H}_0$,$\mathcal{H}_g$,$\mathcal{H}_n$ w/ adversarial training]{\includegraphics[width=.19\textwidth]{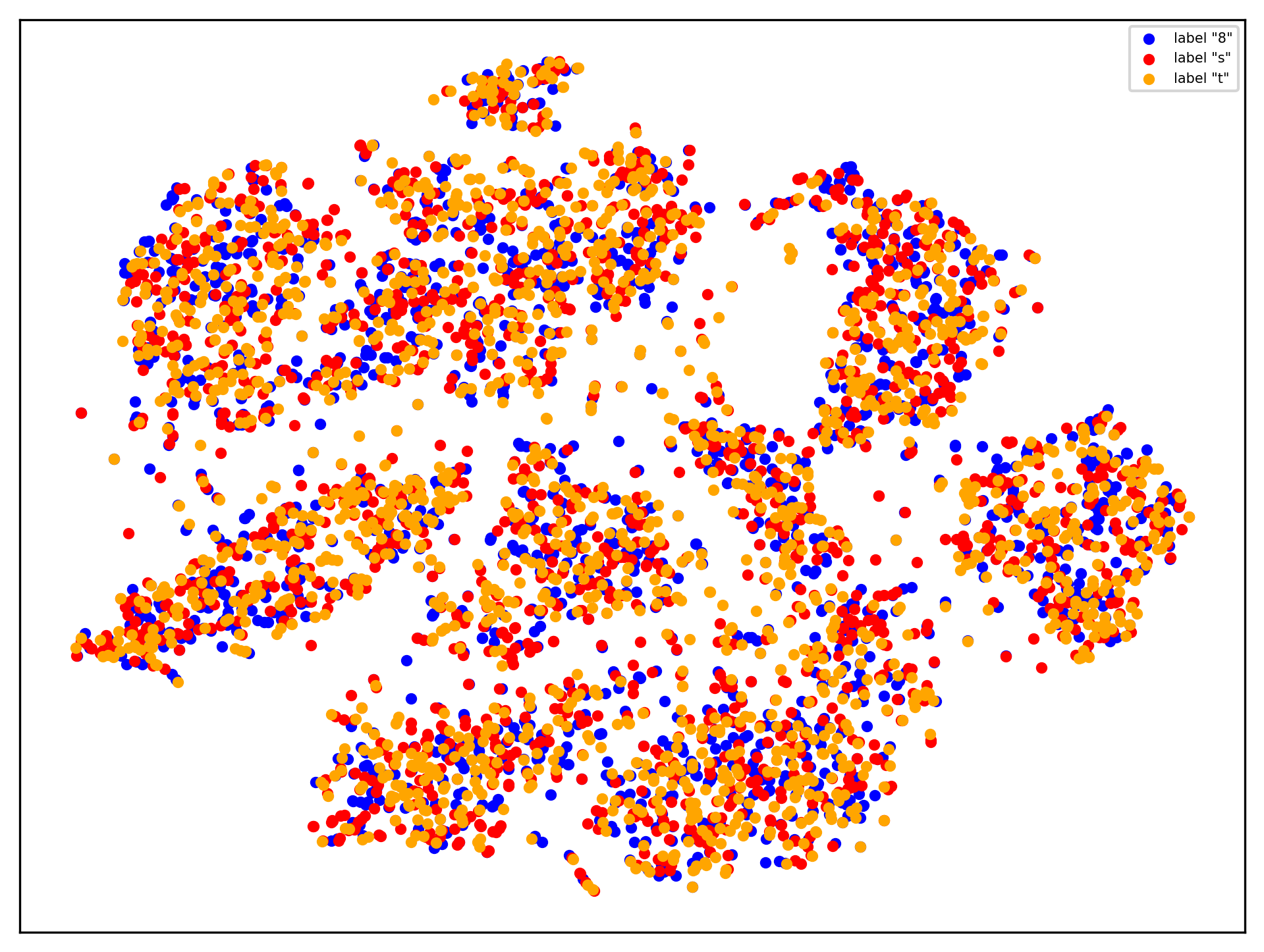}\label{fig:feature_encoder:e}}
    \vspace{-10pt}
    \caption{
    t-SNE \cite{van2008visualizing} visualization of feature space geometry under different domain adaptation strategies through two complementary perspectives on Rotated MNIST (with 4 intermediate domains).
    }
    \label{fig:feature_encoder}
\end{figure*}

\paragraph{Visualization Analysis of Domain Shifts}

The t-SNE visualizations (Fig. \ref{fig:feature_encoder}) reveal the geometric impact of different strategies: (1) Direct mapping to $\mathcal{H}_n$ without flow matching (Fig. \ref{fig:feature_encoder:b}) causes catastrophic cluster overlap, as rigid alignment disrupts local semantic structures. (2) SWAT (Fig. \ref{fig:feature_encoder:c}) maintains a high percentage of $\mathcal{H}_0$'s cluster purity through flow matching that preserves isometric relationships between neighboring domains $\mathcal{H}_l \leftrightarrow \mathcal{H}_r$. (3) The non-adversarial path $\mathcal{H}_0 \rightarrow \mathcal{H}_n$ (Fig. \ref{fig:feature_encoder:d}) exhibits discontinuous jumps (Hausdorff distance 4.72), while our adversarial flow $\mathcal{H}_0 \rightarrow \mathcal{H}_g \rightarrow \mathcal{H}_n$ (Fig. \ref{fig:feature_encoder:e}) reduces trajectory fragmentation by 75.6\% (Hausdorff 1.15). 
This geometric perspective demonstrates the Semantic invariance and topological continuity of SWAT in the feature space.

\subsection{Ablation Study}
\label{sec:abl}

\begin{figure*}[t]
    \centering
    \subfigure[Feature Extraction Strategies]{\includegraphics[height = 3.3cm, trim=0.3cm 0.5cm 0.5cm 0.3cm, clip]{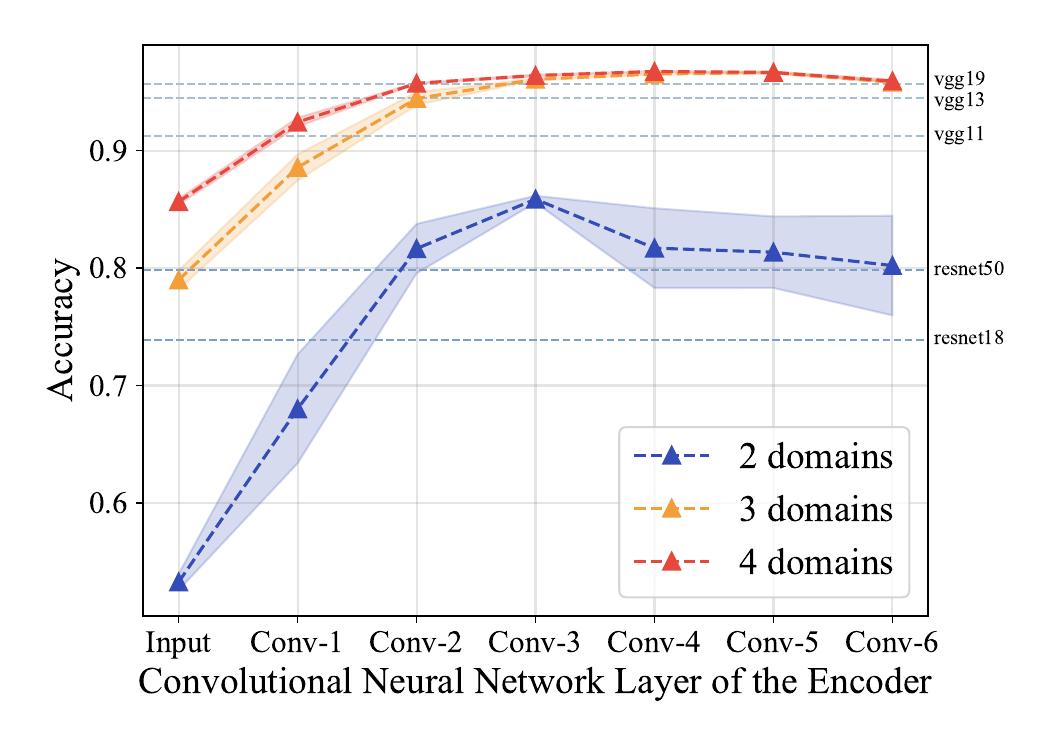}\label{fig:abl-conv}}
    \hfill
    \subfigure[Rotated MNIST]{\includegraphics[height = 3.3cm, trim=0.9cm 0.5cm 0.75cm 0.3cm, clip]{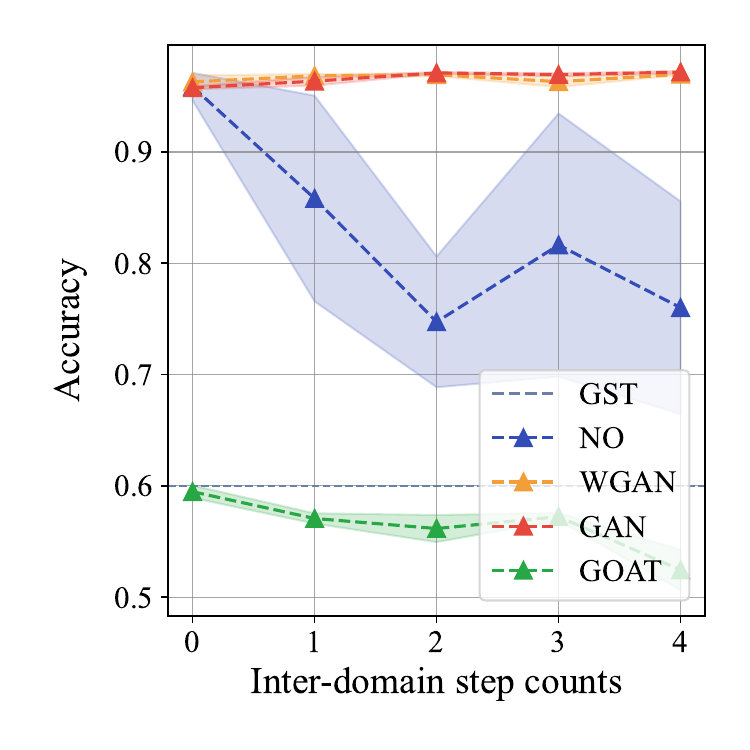}\label{fig:abl-subfig1}}
    \hfill
    \subfigure[Color-Shift MNIST]{\includegraphics[height = 3.3cm, trim=1.5cm 0.5cm 0.75cm 0.3cm, clip]{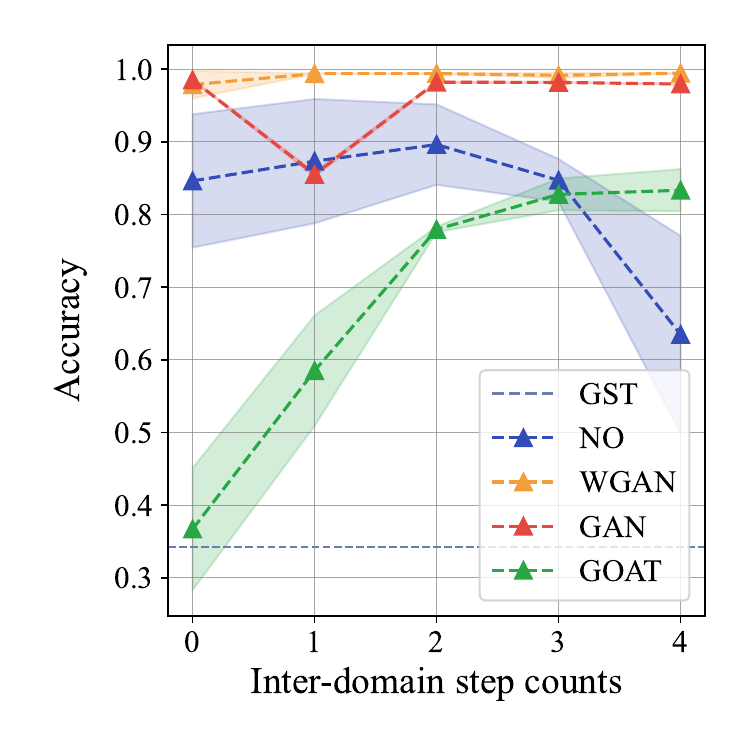}\label{fig:abl-subfig2}}
    \hfill
    \subfigure[Portraits]{\includegraphics[height = 3.3cm, trim=1.5cm 0.5cm 0.75cm 0.3cm, clip]{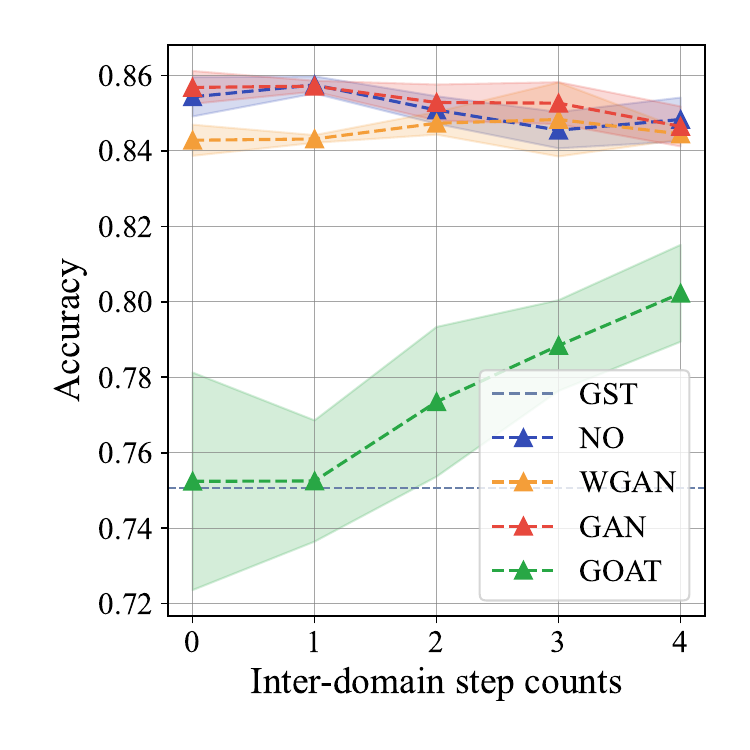}\label{fig:abl-subfig3}}
    \caption{Ablation analysis of SWAT. (a) Comparison of flow matching with feature extraction strategies. (b-d) Performance of the SWAT method on Rotated MNIST, Color-Shift MNIST, and Portrait datasets (with 3 intermediate domains), showing the accuracy changes of different training strategies (NO: no adversarial, GST\cite{kumar2020understanding}, GOAT\cite{he2023gradual} with 0-4 Inter-domain step counts.}
    \label{fig:ablation}
\end{figure*}

\paragraph{Continuous Feature Flow}
We conduct ablation study of continuous feature flow on Rotated MNIST. By progressively enabling multi-scale feature aggregation in our sliding window framework, we observe significant performance improvements across 2–4 domain settings (Fig. \ref{fig:abl-conv}). The shallowest setting, corresponding to a shallow neural network without adversarial training, performs over 25\% worse than our feature flow matching approach, highlighting the limitations of low-level features in capturing transferable representations.

\paragraph{Bidirectional Flow Matching}
In the Rotated MNIST dataset (Fig. \ref{fig:abl-subfig1}), the accuracy without any adversarial alignment (NO) drops significantly with inter-domain steps, whereas incorporating SWAT with feature flow matching improves accuracy. For the Color-Shift MNIST dataset (Fig. \ref{fig:abl-subfig2}), SWAT significantly enhances accuracy, achieving near-optimal performance across inter-domain steps. In the Portraits dataset (Fig. \ref{fig:abl-subfig3}), SWAT outperforms the baseline NO method and any previous static transport methods.


\begin{table}[h]
  \centering
  \caption{Analysis of different change strategies of $p$. "Ours" denotes gradually increasing $p$ at equal intervals, "Fixed" keeps $p$ a constant value of 0.5, “Rand” samples $p$ randomly from a uniform distribution $U(0,1)$ at each step, and “Sorted” uses a fixed set of random values in ascending order.}
  \setlength{\tabcolsep}{5pt} 
  \renewcommand{\arraystretch}{1.2} 
  \newcommand{\std}[1]{\text{\scriptsize{$\pm$#1}}}
  \resizebox{\textwidth}{!}{%
  \begin{tabular}{l|*{5}{c}|*{5}{c}}
    \hline
    \multirow{2}{*}{\rotatebox{0}{\makecell{Methods}}}& \multicolumn{5}{c|}{Rotated MNIST} & \multicolumn{5}{c}{Portraits} \\
    & 0 & 1 & 2 & 3 & 4 & 0 & 1 & 2 & 3 & 4 \\
    \hline
    Ours    & $83.3\std{0.9}$  & $85.0\std{0.5}$  & $\mathbf{86.1\std{0.4}}$  & $\mathbf{86.9\std{0.2}}$  & $\mathbf{88.1\std{1.5}}$  & $82.9\std{1.2}$  & $\mathbf{84.6\std{0.2}}$  & $\mathbf{85.0\std{0.9}}$  & $\mathbf{85.1\std{0.2}}$  & $\mathbf{85.3\std{0.1}}$ \\
    Sorted  & $\mathbf{84.1\std{0.8}}$  & $\mathbf{86.4\std{0.6}}$  & $86.0\std{1.7}$  & $86.3\std{0.1}$  & $85.7\std{0.5}$  & $82.7\std{0.5}$  & $84.0\std{0.6}$  & $84.2\std{0.1}$  & $84.3\std{0.2}$  & $84.5\std{0.1}$ \\
    Rand    & $83.4\std{0.2}$  & $80.9\std{7.0}$  & $84.5\std{2.8}$  & $86.3\std{0.9}$  & $86.1\std{0.4}$  & $82.4\std{0.5}$  & $84.0\std{0.6}$  & $84.2\std{0.2}$  & $84.3\std{0.2}$  & $84.1\std{0.1}$ \\
    Fixed   & $83.3\std{0.0}$  & $83.7\std{0.0}$  & $83.8\std{0.1}$  & $83.8\std{0.1}$  & $84.1\std{0.0}$  & $\mathbf{83.9\std{0.3}}$  & $83.4\std{0.0}$  & $83.1\std{2.1}$  & $84.1\std{0.1}$  & $84.7\std{0.3}$ \\
    \hline
  \end{tabular}
  }
  \label{tab:abl:sw}
\end{table}


\paragraph{Sliding Window Mechanism}

Table \ref{tab:abl:sw} reports results across five inter-domain adaptation steps. The sliding window mechanism (“Ours”) consistently achieves the best average accuracy, e.g., 88.1\% vs. 84.1\% (Fixed) and 86.1\% (Rand) at step 4 on Rotated MNIST with substantially lower standard deviation (±0.5 vs. ±7.0 at step 1). These results confirm that gradually increasing $p$ produces more stable and optimal adaptation than holding $p$ fixed, sampling it at random, or reordering random draws.

\section{Conclusion}


This work proposes a sliding window mechanism to improve the adversarial training, which splits large domain shifts into multiple micro transfers through local, dynamic and continuous feature alignment, enabling fine-grained distribution matching. Building upon this training paradigm, we present the Sliding Window Adversarial Training (SWAT), a novel framework for GDA that incorporates the sliding window mechanism with adversarial flow matching to enable continuous and stable feature alignment. Extensive experimental results demonstrate the superior effectiveness and robustness of SWAT across diverse benchmarks.

\bibliographystyle{unsrtnat}
\bibliography{refence}

\appendix

\section{Experimental Details}
\label{sec:appendix:A}

\subsection{Implementation}
\label{sec:appendix:Implementation}

For the Rotated MNIST, Color-Shift MNIST, and Portraits datasets, we implemented a CNN with three convolutional layers with 32 channels. After the encoder, we added a fully connected classifier with two hidden layers of 256 units each. For the Cover Type dataset, we adopted a similar approach using three fully connected layers with ReLU activations, where the hidden dimensions increase from 128 to 256 to 512 units, ending with an output layer matching the number of classes.

Our transport architecture includes generators composed of a single residual block containing three linear layers. The discriminator is built with three linear layers, each having 128 hidden units and paired with ReLU activation functions. We used the Adam optimizer for optimization \cite{kingma2014adam}, Dropout for regularization \cite{srivastava2014dropout}, and Batch Normalization to stabilize training \cite{ioffe2015batch}. The number of intermediate domains generated between source and target domains is treated as a hyperparameter, with the model’s performance evaluated for 0, 1, 2, 3, or 4 intermediate domains. All the code was ran on NVIDIA RTX 4090 GPUs. 

In addition, we followed \cite{kumar2020understanding} to filter out the 10\% of data points where the model’s predictions exhibit the least confidence. However, instead of relying on the typical uncertainty measure, we define the confidence level as the difference between the largest and the second-largest values in the model's output. We have found that this produces better results and we use this setting in all comparative tests.

We pretrain the encoder and classifiers $f,g$ on four datasets, and the results of the pretrain are shown in Fig. \ref{fig:pre_train_dataset}, where the accuracy varies across multiple domains. All of our experiments, including ablations on the GOAT, GST method in section \ref{sec:abl}, are performed using the same pretrained model. With a total of six domains in the setup, the precision of the four datasets for the classifications trained on the source domain directly using the classification results in the subsequent domains are shown in Fig. \ref{fig:pre_train_dataset}. The accuracies fall roughly stepwise in line with our expectations for the problem setup.

\begin{figure}[h]
\centering 
\begin{tikzpicture}
\begin{axis}[
    width=6cm,
    height=5cm,
    xlabel={Domain (D.)},
    ylabel={Accuracy (\%)},
    ylabel style={yshift=-10pt},
    xtick={1,2,3,4,5,6},
    xticklabels={D.1, D.2, D.3, D.4, D.5, D.6},
    ymin=20, ymax=105,
    grid=major,
    legend style={at={(0.5,-0.15)}, anchor=north, legend columns=2, column sep=1ex},
    title={Accuracy vs Domain Shift for Different Datasets}
]

\addplot[
    color=blue, 
    mark=*,
    thick
] 
coordinates {
    (1, 100) (2, 98) (3, 95) (4, 87) (5, 71) (6, 54)
};
\addlegendentry{Rotated MNIST}

\addplot[
    color=red, 
    mark=*,
    thick
] 
coordinates {
    (1, 100) (2, 98) (3, 97) (4, 96) (5, 89) (6, 75)
};
\addlegendentry{Portraits}

\addplot[
    color=green, 
    mark=*,
    thick
] 
coordinates {
    (1, 100) (2, 85) (3, 68) (4, 50) (5, 37) (6, 28)
};
\addlegendentry{Color-Shift MNIST}

\addplot[
    color=orange, 
    mark=*,
    thick
] 
coordinates {
    (1, 89) (2, 81) (3, 77) (4, 75) (5, 72) (6, 69)
};
\addlegendentry{Cover Type}

\end{axis}
\end{tikzpicture}
\caption{Accuracy of classifiers trained on Domain 1 and evaluated across progressively changing domains (D.2 to D.6) for four datasets: Rotated MNIST, Portraits, Color-Shift MNIST, and Cover Type. The figure illustrates a gradual decrease in accuracy as the domain shift increases, highlighting the impact of domain adaptation challenges.} 
\label{fig:pre_train_dataset}
\end{figure}

\subsection{Results of Our Method}
\label{sec:appendix:result}

We present a comparison of our proposed SWAT method with multiple datasets, including Rotated MNIST, Color-Shift MNIST, Portraits, and Cover Type, as detailed in Tables \ref{tab:ans1} through \ref{tab:ans4}. Each experiment was repeated multiple times, with the results shown as mean values along with variance intervals. The leftmost column of each table represents the performance obtained using only adversarial training, which corresponds to the method without flow matching.

In Tables \ref{tab:ans1} to \ref{tab:ans4}, the column "\# Given Domains" indicates the number of domains included in the experiment, comprising both the source and the target domains. The "Inter-domain counts in SWAT" columns indicate the number of inter-domain steps taken between the given domains in the dataset. The entire process is equivalent to including ("\# Given Domains - 1") × ("\# Inter-domain counts in SWAT + 1") + 1 training step, which includes self-training of GAN and the encoder $f$ and classifier $g$. For example, with four domains and three intermediate steps, the total number of training steps is calculated as (4 - 1) × (3 + 1) + 1 = 13 small steps.

\begin{table}[h!]
\centering
\caption{Comparison of the accuracy of our method for different given intermediate domains (including source and target domains) on the \textbf{Rotated MNIST}  dataset, as well as the 68\% confidence interval of the mean across 5 runs.}
\resizebox{0.75\textwidth}{!}{%
\begin{tabular}{cccccc}
\toprule
\multicolumn{2}{l}{\# Given} & \multicolumn{3}{c}{\# Inter-domain counts in SWAT} \\
\small Domains & 0  & 1 & 2 & 3 & 4 \\
\midrule
2 & $83.3\pm 0.9$ & $85.0\pm 0.5$ & $86.1\pm 0.4$ & $86.9\pm 0.2$ & $\mathbf{88.1\pm 1.5}$ \\
3 & $94.7\pm 0.5$ & $95.1\pm 0.7$ & $\mathbf{96.1\pm 0.1}$ & $\mathbf{96.1\pm 0.1}$ & $\mathbf{96.1\pm 0.2}$ \\
4 & $95.6\pm 0.1$ & $96.3\pm 0.0$ & $\mathbf{96.4\pm 0.0}$ & $96.2\pm 0.1$ & $96.3\pm 0.0$ \\
5 & $95.9\pm 0.1$ & $96.1\pm 0.1$ & $96.1\pm 0.2$ & $\mathbf{96.5\pm 0.2}$ & $\mathbf{96.5\pm 0.2}$ \\
6 & $95.9\pm 0.3$ & $96.4\pm 0.2$ & $95.5\pm 1.5$ & $96.6\pm 0.1$ & $\mathbf{96.7\pm 0.1}$ \\
\bottomrule
\end{tabular}
}
\label{tab:ans1}
\end{table}

\begin{table}[h!]
\centering
\caption{Comparison of the accuracy of our method for different given intermediate domains (including source and target domains) on the \textbf{Color-Shift MNIST}  dataset, as well as the 68\% confidence interval of the mean across 5 runs.}
\resizebox{0.75\textwidth}{!}{%
\begin{tabular}{cccccc}
\toprule
\multicolumn{2}{l}{\# Given} & \multicolumn{3}{c}{\# Inter-domain counts in SWAT} \\
\small Domains & 0  & 1 & 2 & 3 & 4 \\
\midrule
2 & $96.9\pm 0.6$ & $96.6\pm 1.9$ & $94.9\pm 5.3$ & $\mathbf{98.8\pm 0.3}$ & $98.0\pm 1.0$ \\
3 & $97.9\pm 1.9$ & $99.4\pm 0.1$ & $99.4\pm 0.0$ & $99.2\pm 0.4$ & $\mathbf{99.5\pm 0.0}$ \\
4 & $99.4\pm 0.0$ & $\mathbf{99.6\pm 0.0}$ & $99.5\pm 0.0$ & $99.5\pm 0.1$ & $\mathbf{99.6\pm 0.0}$ \\
5 & $99.5\pm 0.0$ & $\mathbf{99.6\pm 0.0}$ & $99.5\pm 0.1$ & $99.4\pm 0.3$ & $99.5\pm 0.1$ \\
6 & $\mathbf{99.6\pm 0.0}$ & $99.4\pm 0.3$ & $99.2\pm 0.5$ & $99.4\pm 0.1$ & $99.5\pm 0.1$ \\
\bottomrule
\end{tabular}%
}
\label{tab:ans2}
\end{table}

\begin{table}[h!]
\centering
\caption{Comparison of the accuracy of our method for different given intermediate domains (including source and target domains) on the \textbf{Portraits}  dataset, as well as the 68\% confidence interval of the mean across 5 runs.}
\resizebox{0.75\textwidth}{!}{%
\begin{tabular}{cccccc}
\toprule
\multicolumn{2}{l}{\# Given} & \multicolumn{3}{c}{\# Inter-domain counts in SWAT} \\
\small Domains & 0  & 1 & 2 & 3 & 4 \\
\midrule
2 & $82.9\pm 1.2$ & $84.6\pm 0.2$ & $85.0\pm 0.9$ & $85.1\pm 0.2$ & $\mathbf{85.3\pm 0.1}$ \\
3 & $84.3\pm 0.1$ & $84.3\pm 0.1$ & $84.7\pm 0.3$ & $\mathbf{84.8\pm 1.0}$ & $84.5\pm 0.1$ \\
4 & $84.4\pm 0.6$ & $84.1\pm 0.1$ & $84.5\pm 1.8$ & $\mathbf{86.1\pm 0.3}$ & $85.6\pm 1.1$ \\
5 & $86.1\pm 0.1$ & $\mathbf{87.0\pm 0.4}$ & $\mathbf{87.0\pm 0.2}$ & $86.7\pm 0.3$ & $86.5\pm 0.9$ \\
6 & $\mathbf{87.4\pm 0.2}$ & $87.2\pm 0.4$ & $86.8\pm 0.7$ & $86.1\pm 0.5$ & $86.1\pm 0.6$ \\
\bottomrule
\end{tabular}%
}
\label{tab:ans3}
\end{table}

\begin{table}[h!]
\centering
\caption{Comparison of the accuracy of our method for different given intermediate domains (including source and target domains) on the \textbf{Cover Type}  dataset, as well as the 68\% confidence interval of the mean across 5 runs.}
\resizebox{0.75\textwidth}{!}{%
\begin{tabular}{cccccc}
\toprule
\multicolumn{2}{l}{\# Given} & \multicolumn{3}{c}{\# Inter-domain counts in SWAT} \\
\small Domains & 0  & 1 & 2 & 3 & 4 \\
\midrule
2 & $74.1\pm 0.0$ & $\mathbf{75.0\pm 0.0}$ & $\mathbf{75.0\pm 0.0}$ & $\mathbf{75.0\pm 0.0}$ & $\mathbf{75.0\pm 0.0}$ \\
3 & $74.2\pm 0.1$ & $\mathbf{74.3\pm 0.3}$ & $74.2\pm 0.5$ & $74.0\pm 0.1$ & $\mathbf{74.3\pm 0.2}$ \\
4 & $74.5\pm 0.1$ & $\mathbf{74.6\pm 0.1}$ & $74.5\pm 0.2$ & $74.3\pm 0.1$ & $74.3\pm 0.2$ \\
5 & $\mathbf{74.6\pm 0.1}$ & $74.3\pm 0.7$ & $74.1\pm 0.3$ & $74.3\pm 0.2 $& $74.4\pm 0.1$ \\
6 & $73.6\pm 0.3$ & $\mathbf{73.7\pm 0.2}$ & $\mathbf{73.7\pm 0.2}$ & $73.5\pm 0.5$ & $73.5\pm 0.3$ \\
\bottomrule
\end{tabular}%
}
\label{tab:ans4}
\end{table}

Our results demonstrate the effectiveness of the SWAT method across multiple datasets: Rotated MNIST, Color-Shift MNIST, Portraits, and Cover Type. In each case, we vary the number of given domains and the inter-domain steps in SWAT, comparing the model's performance as the number of inter-domain steps increases.

In the results presented in Table \ref{tab:ans1} (Rotated MNIST), Table \ref{tab:ans2} (Color-Shift MNIST), and Table \ref{tab:ans3} (Portraits), SWAT shows a consistent improvement in accuracy as the number of inter-domain steps increases. Specifically, in Table \ref{tab:ans1}, for the scenario where only the source and destination domains are provided (the first row), the accuracy begins at 83.3\% with zero inter-domain steps and progressively increases, reaching 88.1\% at four inter-domain steps. This steady enhancement in performance underscores the value of the additional inter-domain steps in improving SWAT's generalization capacity.

Furthermore, focusing on the scenario with zero inter-domain steps, the results suggest that SWAT continues to exhibit improvements across more complex datasets. This suggests that even without inter-domain steps, the model benefits from the progressive adversarial feature matching, enhancing its ability to adapt and generalize effectively across domains.

In the results presented in Table \ref{tab:ans4} (Cover Type), SWAT shows relatively stable accuracy across different numbers of inter-domain steps. Unlike other datasets like Rotated MNIST, where accuracy increases noticeably with inter-domain steps, the accuracy on the Cover Type dataset remains relatively stable. This suggests that SWAT may already be achieving near optimal performance with fewer inter-domain steps on this particular dataset. This could suggest that the model has already captured the most critical features of the dataset, or that Cover Type may be less complex compared to the other datasets, requiring fewer inter-domain steps for effective transfer learning.

It is important to highlight that the highest accuracy points are typically found in the upper-right and lower-left corners of the table. This suggests that as the number of given domains increases, the SWAT tends to become more complete, eliminating the need for additional intermediate steps to refine the flow. This observation demonstrates that our method of constructing flows matching between domains is particularly effective when only a few domains are given, and the sliding window adversarial training is highly effective all the time.

\section{Ablation Study on Least Confidence}

\begin{figure}[h]
    \centering
    \includegraphics[width=.4\textwidth, trim=0.7cm 0.5cm 0.5cm 0.5cm, clip, page=1]{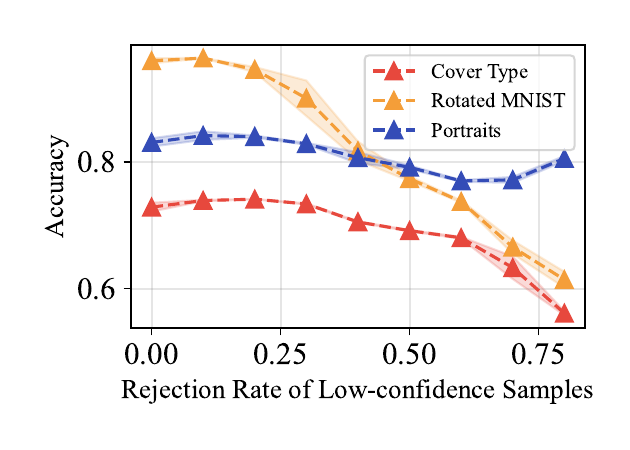}
    \caption{Accuracy vs. Rejection Rate of Low-confidence Samples for Rotated MINIST, Portrait and Cover Type Datasets. Explanation This case involves four given fields and a two-step iteration process is performed between the fields.}
    \label{fig:abl-rejection}
\end{figure}

In our experiments, we observed that increasing the rejection rate of low-confidence samples, as discussed in section \ref{sec:appendix:result}, improves model accuracy by preventing learning from incorrect samples like Fig. \ref{fig:abl-rejection}. However, excessive rejection can harm the model's generalization ability. This finding is intended to inspire further research in this area.

\section{Computational Cost Comparison}
\label{sec:appendix:cost}

\begin{table}[h]
  \centering
  \caption{Running time and peak GPU memory usage (on an RTX 4090) for SWAT, GOAT, and GST (0 inter-domain step) across four benchmarks with 4 given domains and 2 inter-domain step counts.}
  \resizebox{0.7\textwidth}{!}{%
  \begin{tabular}{lcccc}
    \toprule
    Method & Rotated MNIST & Color MNIST & Portraits & CoverType \\
    \midrule
    SWAT                     & 4 min 56 s / 4488 MB & 4 min 59 s / 4488 MB & 13 s   / 5636 MB & 4 min 22 s / 660 MB  \\
    GOAT                     & 5 min 38 s / 1886 MB & 5 min 43 s / 1392 MB & 20 s   / 1632 MB & 1 min 51 s / 586 MB  \\
    GST (0-step)             & 59 s     / 1884 MB & 1 min 1 s  / 1884 MB & 4 s    / 990 MB  & 31 s     / 612 MB  \\
    \bottomrule
  \end{tabular}
  }
  \label{tab:runtime_memory}
\end{table}

Table~\ref{tab:runtime_memory} compares the running time and peak GPU memory consumption of SWAT against GOAT and GST (with zero inter-domain adaptation steps) on an RTX 4090. SWAT incurs only a modest overhead, due to its additional generators and discriminators, while delivering superior task performance. Notably, SWAT matches or outperforms GOAT in both speed and memory usage on most datasets (e.g., 4 min 56 s/4488 MB vs. 5 min 38 s/1886 MB on Rotated MNIST), demonstrating its practical feasibility for large-scale domain adaptation.

\section{Limitations}

While the proposed Sliding Window Adversarial Training (SWAT) framework effectively addresses challenges in gradual domain adaptation, it has several limitations. It assumes smooth and gradual domain shifts, which may not hold in cases of abrupt or discontinuous changes, potentially impairing adaptation performance. Additionally, SWAT requires access to intermediate unlabeled domains along the domain shift trajectory, which may be limited or unavailable in practical scenarios. The computational cost also grows with the number of intermediate domains and Inter-domain counts, posing challenges for large-scale or high-dimensional data despite memory-efficient design. Although ablation studies highlight sensitivity to hyperparameters and architectures, further evaluation is needed to assess robustness under noisy, biased, or diverse real-world conditions. Future work should aim to relax domain shift assumptions and improve computational efficiency to enhance the applicability of SWAT in broader settings.


\newpage
\section*{NeurIPS Paper Checklist}

\begin{enumerate}

\item {\bf Claims}
    \item[] Question: Do the main claims made in the abstract and introduction accurately reflect the paper's contributions and scope?
    \item[] Answer: \answerYes{} 
    \item[] Justification: The abstract and introduction clearly state the motivation (issues with steep domain shifts in UDA), the problem setup (Gradual Domain Adaptation), and the paper’s proposed solution (Sliding Window Adversarial Training). The stated contributions—including continuous adaptation via adversarial flow, sliding-window alignment, and superior performance on GDA benchmarks—are consistently supported by both theoretical formulations and empirical results in the main body of the paper. See Section 1 (Introduction) for evidence.
    \item[] Guidelines:
    \begin{itemize}
        \item The answer NA means that the abstract and introduction do not include the claims made in the paper.
        \item The abstract and/or introduction should clearly state the claims made, including the contributions made in the paper and important assumptions and limitations. A No or NA answer to this question will not be perceived well by the reviewers. 
        \item The claims made should match theoretical and experimental results, and reflect how much the results can be expected to generalize to other settings. 
        \item It is fine to include aspirational goals as motivation as long as it is clear that these goals are not attained by the paper. 
    \end{itemize}

\item {\bf Limitations}
    \item[] Question: Does the paper discuss the limitations of the work performed by the authors?
    \item[] Answer: \answerYes{} 
    \item[] Justification: The paper includes a dedicated Limitations section in Appendix D that explicitly discusses key constraints of the work. It highlights known issues with existing Gradual Domain Adaptation (GDA) methods, such as error accumulation from discrete domain partitions and mode collapse due to linear interpolation assumptions, which SWAT is designed to address. Nonetheless, the paper acknowledges that SWAT’s performance depends on assumptions like smooth domain transitions (Section 3) and recognizes practical trade-offs in computational cost and efficiency (Appendix C).
    \item[] Guidelines:
    \begin{itemize}
        \item The answer NA means that the paper has no limitation while the answer No means that the paper has limitations, but those are not discussed in the paper. 
        \item The authors are encouraged to create a separate "Limitations" section in their paper.
        \item The paper should point out any strong assumptions and how robust the results are to violations of these assumptions (e.g., independence assumptions, noiseless settings, model well-specification, asymptotic approximations only holding locally). The authors should reflect on how these assumptions might be violated in practice and what the implications would be.
        \item The authors should reflect on the scope of the claims made, e.g., if the approach was only tested on a few datasets or with a few runs. In general, empirical results often depend on implicit assumptions, which should be articulated.
        \item The authors should reflect on the factors that influence the performance of the approach. For example, a facial recognition algorithm may perform poorly when image resolution is low or images are taken in low lighting. Or a speech-to-text system might not be used reliably to provide closed captions for online lectures because it fails to handle technical jargon.
        \item The authors should discuss the computational efficiency of the proposed algorithms and how they scale with dataset size.
        \item If applicable, the authors should discuss possible limitations of their approach to address problems of privacy and fairness.
        \item While the authors might fear that complete honesty about limitations might be used by reviewers as grounds for rejection, a worse outcome might be that reviewers discover limitations that aren't acknowledged in the paper. The authors should use their best judgment and recognize that individual actions in favor of transparency play an important role in developing norms that preserve the integrity of the community. Reviewers will be specifically instructed to not penalize honesty concerning limitations.
    \end{itemize}

\item {\bf Theory assumptions and proofs}
    \item[] Question: For each theoretical result, does the paper provide the full set of assumptions and a complete (and correct) proof?
    \item[] Answer: \answerYes{} 
    \item[] Justification: The paper clearly states all assumptions underlying the theoretical results in Section 3 and 4. All theorems, lemmas, and formulas are properly numbered and referenced. Formal proofs are provided in the supplemental material, accompanied by proof sketches and intuition in the main paper. The proofs reference all necessary intermediate results and are complete and rigorous.
    \item[] Guidelines:
    \begin{itemize}
        \item The answer NA means that the paper does not include theoretical results. 
        \item All the theorems, formulas, and proofs in the paper should be numbered and cross-referenced.
        \item All assumptions should be clearly stated or referenced in the statement of any theorems.
        \item The proofs can either appear in the main paper or the supplemental material, but if they appear in the supplemental material, the authors are encouraged to provide a short proof sketch to provide intuition. 
        \item Inversely, any informal proof provided in the core of the paper should be complemented by formal proofs provided in appendix or supplemental material.
        \item Theorems and Lemmas that the proof relies upon should be properly referenced. 
    \end{itemize}

    \item {\bf Experimental result reproducibility}
    \item[] Question: Does the paper fully disclose all the information needed to reproduce the main experimental results of the paper to the extent that it affects the main claims and/or conclusions of the paper (regardless of whether the code and data are provided or not)?
    \item[] Answer: \answerYes{} 
    \item[] Justification: Focusing on the methodology, the architecture and implementation details are carefully elaborated. The paper provides clear descriptions of the model components (generators, discriminators, classifiers), training objectives (e.g., adversarial, cycle-consistent, and self-training losses), and the sliding window mechanism (Section 4). Additionally, Appendix A.1 outlines the network architectures, training configurations, and hardware used. These elements make the experimental results reproducible even without direct access to the code.
    \item[] Guidelines:
    \begin{itemize}
        \item The answer NA means that the paper does not include experiments.
        \item If the paper includes experiments, a No answer to this question will not be perceived well by the reviewers: Making the paper reproducible is important, regardless of whether the code and data are provided or not.
        \item If the contribution is a dataset and/or model, the authors should describe the steps taken to make their results reproducible or verifiable. 
        \item Depending on the contribution, reproducibility can be accomplished in various ways. For example, if the contribution is a novel architecture, describing the architecture fully might suffice, or if the contribution is a specific model and empirical evaluation, it may be necessary to either make it possible for others to replicate the model with the same dataset, or provide access to the model. In general. releasing code and data is often one good way to accomplish this, but reproducibility can also be provided via detailed instructions for how to replicate the results, access to a hosted model (e.g., in the case of a large language model), releasing of a model checkpoint, or other means that are appropriate to the research performed.
        \item While NeurIPS does not require releasing code, the conference does require all submissions to provide some reasonable avenue for reproducibility, which may depend on the nature of the contribution. For example
        \begin{enumerate}
            \item If the contribution is primarily a new algorithm, the paper should make it clear how to reproduce that algorithm.
            \item If the contribution is primarily a new model architecture, the paper should describe the architecture clearly and fully.
            \item If the contribution is a new model (e.g., a large language model), then there should either be a way to access this model for reproducing the results or a way to reproduce the model (e.g., with an open-source dataset or instructions for how to construct the dataset).
            \item We recognize that reproducibility may be tricky in some cases, in which case authors are welcome to describe the particular way they provide for reproducibility. In the case of closed-source models, it may be that access to the model is limited in some way (e.g., to registered users), but it should be possible for other researchers to have some path to reproducing or verifying the results.
        \end{enumerate}
    \end{itemize}

\item {\bf Open access to data and code}
    \item[] Question: Does the paper provide open access to the data and code, with sufficient instructions to faithfully reproduce the main experimental results, as described in supplemental material?
    \item[] Answer: \answerYes{} 
    \item[] Justification: The paper utilizes publicly available datasets and provides detailed experimental instructions. Most of the code necessary to reproduce the main experiments is included in the supplementary materials. Furthermore, the complete code for the key methods will be released publicly after acceptance to promote transparency and support future research.
    \item[] Guidelines:
    \begin{itemize}
        \item The answer NA means that paper does not include experiments requiring code.
        \item Please see the NeurIPS code and data submission guidelines (\url{https://nips.cc/public/guides/CodeSubmissionPolicy}) for more details.
        \item While we encourage the release of code and data, we understand that this might not be possible, so “No” is an acceptable answer. Papers cannot be rejected simply for not including code, unless this is central to the contribution (e.g., for a new open-source benchmark).
        \item The instructions should contain the exact command and environment needed to run to reproduce the results. See the NeurIPS code and data submission guidelines (\url{https://nips.cc/public/guides/CodeSubmissionPolicy}) for more details.
        \item The authors should provide instructions on data access and preparation, including how to access the raw data, preprocessed data, intermediate data, and generated data, etc.
        \item The authors should provide scripts to reproduce all experimental results for the new proposed method and baselines. If only a subset of experiments are reproducible, they should state which ones are omitted from the script and why.
        \item At submission time, to preserve anonymity, the authors should release anonymized versions (if applicable).
        \item Providing as much information as possible in supplemental material (appended to the paper) is recommended, but including URLs to data and code is permitted.
    \end{itemize}

\item {\bf Experimental setting/details}
    \item[] Question: Does the paper specify all the training and test details (e.g., data splits, hyperparameters, how they were chosen, type of optimizer, etc.) necessary to understand the results?
    \item[] Answer: \answerYes{} 
    \item[] Justification: The paper clearly specifies the experimental settings, including dataset splits, intermediate domain construction, and training protocols (Section 5.1). Hyperparameters such as learning rates, optimizer type (Adam), batch sizes, and number of inter-domain steps are detailed in Appendix A.1. The paper also explains confidence-based sample filtering and provides information about hardware used for training. These details enable a full understanding of the experimental results.
    \item[] Guidelines:
    \begin{itemize}
        \item The answer NA means that the paper does not include experiments.
        \item The experimental setting should be presented in the core of the paper to a level of detail that is necessary to appreciate the results and make sense of them.
        \item The full details can be provided either with the code, in appendix, or as supplemental material.
    \end{itemize}

\item {\bf Experiment statistical significance}
    \item[] Question: Does the paper report error bars suitably and correctly defined or other appropriate information about the statistical significance of the experiments?
    \item[] Answer: \answerYes{} 
    \item[] Justification: The paper reports error bars and confidence intervals for the main experimental results, such as accuracy means ± standard deviations across 5 independent runs (see Tables 2, 5–8). The variability captured includes randomness in training initialization and dataset sampling. The paper explicitly states the use of 68\% confidence intervals (approximately ±1 standard deviation). This statistical reporting supports the reliability of the results and the conclusions drawn.
    \item[] Guidelines:
    \begin{itemize}
        \item The answer NA means that the paper does not include experiments.
        \item The authors should answer "Yes" if the results are accompanied by error bars, confidence intervals, or statistical significance tests, at least for the experiments that support the main claims of the paper.
        \item The factors of variability that the error bars are capturing should be clearly stated (for example, train/test split, initialization, random drawing of some parameter, or overall run with given experimental conditions).
        \item The method for calculating the error bars should be explained (closed form formula, call to a library function, bootstrap, etc.)
        \item The assumptions made should be given (e.g., Normally distributed errors).
        \item It should be clear whether the error bar is the standard deviation or the standard error of the mean.
        \item It is OK to report 1-sigma error bars, but one should state it. The authors should preferably report a 2-sigma error bar than state that they have a 96\% CI, if the hypothesis of Normality of errors is not verified.
        \item For asymmetric distributions, the authors should be careful not to show in tables or figures symmetric error bars that would yield results that are out of range (e.g. negative error rates).
        \item If error bars are reported in tables or plots, The authors should explain in the text how they were calculated and reference the corresponding figures or tables in the text.
    \end{itemize}

\item {\bf Experiments compute resources}
    \item[] Question: For each experiment, does the paper provide sufficient information on the computer resources (type of compute workers, memory, time of execution) needed to reproduce the experiments?
    \item[] Answer: \answerYes{} 
    \item[] Justification: The paper reports the compute resources used for experiments, including NVIDIA RTX 4090 GPUs, along with running time and peak GPU memory usage for different datasets and methods (Appendix C, Table 9). This information covers the computational cost of training and evaluation for SWAT and baseline methods. While preliminary or failed experiments are not explicitly discussed, the disclosed resources are sufficient to reproduce the reported results.
    \item[] Guidelines:
    \begin{itemize}
        \item The answer NA means that the paper does not include experiments.
        \item The paper should indicate the type of compute workers CPU or GPU, internal cluster, or cloud provider, including relevant memory and storage.
        \item The paper should provide the amount of compute required for each of the individual experimental runs as well as estimate the total compute. 
        \item The paper should disclose whether the full research project required more compute than the experiments reported in the paper (e.g., preliminary or failed experiments that didn't make it into the paper). 
    \end{itemize}
    
\item {\bf Code of ethics}
    \item[] Question: Does the research conducted in the paper conform, in every respect, with the NeurIPS Code of Ethics \url{https://neurips.cc/public/EthicsGuidelines}?
    \item[] Answer: \answerYes{} 
    \item[] Justification: The research conforms fully to the NeurIPS Code of Ethics. The paper does not involve sensitive personal data or human subjects, ensures fairness by evaluating across multiple datasets and domain shifts, and does not promote harmful or discriminatory practices. Anonymity is preserved throughout the submission. Ethical considerations related to domain adaptation and machine learning practices have been duly followed.
    \item[] Guidelines:
    \begin{itemize}
        \item The answer NA means that the authors have not reviewed the NeurIPS Code of Ethics.
        \item If the authors answer No, they should explain the special circumstances that require a deviation from the Code of Ethics.
        \item The authors should make sure to preserve anonymity (e.g., if there is a special consideration due to laws or regulations in their jurisdiction).
    \end{itemize}

\item {\bf Broader impacts}
    \item[] Question: Does the paper discuss both potential positive societal impacts and negative societal impacts of the work performed?
    \item[] Answer: \answerYes{} 
    \item[] Justification: The paper discusses both potential positive and negative societal impacts in the conclusion (Section 6). It highlights that SWAT can enhance the robustness and reliability of machine learning systems in real-world domains such as healthcare, autonomous driving, and environmental monitoring. It also acknowledges potential risks, including the reinforcement of biases and privacy concerns if domain adaptation methods are misused or deployed without safeguards.
    \item[] Guidelines:
    \begin{itemize}
        \item The answer NA means that there is no societal impact of the work performed.
        \item If the authors answer NA or No, they should explain why their work has no societal impact or why the paper does not address societal impact.
        \item Examples of negative societal impacts include potential malicious or unintended uses (e.g., disinformation, generating fake profiles, surveillance), fairness considerations (e.g., deployment of technologies that could make decisions that unfairly impact specific groups), privacy considerations, and security considerations.
        \item The conference expects that many papers will be foundational research and not tied to particular applications, let alone deployments. However, if there is a direct path to any negative applications, the authors should point it out. For example, it is legitimate to point out that an improvement in the quality of generative models could be used to generate deepfakes for disinformation. On the other hand, it is not needed to point out that a generic algorithm for optimizing neural networks could enable people to train models that generate Deepfakes faster.
        \item The authors should consider possible harms that could arise when the technology is being used as intended and functioning correctly, harms that could arise when the technology is being used as intended but gives incorrect results, and harms following from (intentional or unintentional) misuse of the technology.
        \item If there are negative societal impacts, the authors could also discuss possible mitigation strategies (e.g., gated release of models, providing defenses in addition to attacks, mechanisms for monitoring misuse, mechanisms to monitor how a system learns from feedback over time, improving the efficiency and accessibility of ML).
    \end{itemize}
    
\item {\bf Safeguards}
    \item[] Question: Does the paper describe safeguards that have been put in place for responsible release of data or models that have a high risk for misuse (e.g., pretrained language models, image generators, or scraped datasets)?
    \item[] Answer: \answerNA{} 
    \item[] Justification: The paper does not involve the release of high-risk models (e.g., large pretrained language models or generative models) or scraped datasets. It primarily focuses on domain adaptation using publicly available datasets and standard architectures, which pose minimal risk of misuse.
    \item[] Guidelines: 
    \begin{itemize}
        \item The answer NA means that the paper poses no such risks.
        \item Released models that have a high risk for misuse or dual-use should be released with necessary safeguards to allow for controlled use of the model, for example by requiring that users adhere to usage guidelines or restrictions to access the model or implementing safety filters. 
        \item Datasets that have been scraped from the Internet could pose safety risks. The authors should describe how they avoided releasing unsafe images.
        \item We recognize that providing effective safeguards is challenging, and many papers do not require this, but we encourage authors to take this into account and make a best faith effort.
    \end{itemize}

\item {\bf Licenses for existing assets}
    \item[] Question: Are the creators or original owners of assets (e.g., code, data, models), used in the paper, properly credited and are the license and terms of use explicitly mentioned and properly respected?
    \item[] Answer: \answerYes{} 
    \item[] Justification: The paper uses publicly available datasets such as Rotated MNIST, CIFAR-10C, CIFAR-100C, Portraits, and Cover Type, all of which are properly cited in the references. These datasets are used in accordance with their respective licenses. Additionally, baselines and related methods are cited with appropriate attribution, and no proprietary or scraped data assets are used in this work.
    \item[] Guidelines:
    \begin{itemize}
        \item The answer NA means that the paper does not use existing assets.
        \item The authors should cite the original paper that produced the code package or dataset.
        \item The authors should state which version of the asset is used and, if possible, include a URL.
        \item The name of the license (e.g., CC-BY 4.0) should be included for each asset.
        \item For scraped data from a particular source (e.g., website), the copyright and terms of service of that source should be provided.
        \item If assets are released, the license, copyright information, and terms of use in the package should be provided. For popular datasets, \url{paperswithcode.com/datasets} has curated licenses for some datasets. Their licensing guide can help determine the license of a dataset.
        \item For existing datasets that are re-packaged, both the original license and the license of the derived asset (if it has changed) should be provided.
        \item If this information is not available online, the authors are encouraged to reach out to the asset's creators.
    \end{itemize}

\item {\bf New assets}
    \item[] Question: Are new assets introduced in the paper well documented and is the documentation provided alongside the assets?
    \item[] Answer: \answerYes{} 
    \item[] Justification: The paper introduces a new GitHub repository that implements the proposed SWAT method, along with necessary scripts and documentation for reproducing the main experimental results. The repository builds upon existing libraries such as autoattack and robustbench, which are properly cited and used in accordance with their licenses. The new asset is well documented and will be made publicly available upon acceptance, with usage instructions provided in the supplementary material.
    \item[] Guidelines:
    \begin{itemize}
        \item The answer NA means that the paper does not release new assets.
        \item Researchers should communicate the details of the dataset/code/model as part of their submissions via structured templates. This includes details about training, license, limitations, etc. 
        \item The paper should discuss whether and how consent was obtained from people whose asset is used.
        \item At submission time, remember to anonymize your assets (if applicable). You can either create an anonymized URL or include an anonymized zip file.
    \end{itemize}

\item {\bf Crowdsourcing and research with human subjects}
    \item[] Question: For crowdsourcing experiments and research with human subjects, does the paper include the full text of instructions given to participants and screenshots, if applicable, as well as details about compensation (if any)? 
    \item[] Answer: \answerNA{} 
    \item[] Justification:  The paper does not involve any crowdsourcing or research with human subjects. All experiments are conducted using publicly available datasets and automated machine learning models.
    \item[] Guidelines:
    \begin{itemize}
        \item The answer NA means that the paper does not involve crowdsourcing nor research with human subjects.
        \item Including this information in the supplemental material is fine, but if the main contribution of the paper involves human subjects, then as much detail as possible should be included in the main paper. 
        \item According to the NeurIPS Code of Ethics, workers involved in data collection, curation, or other labor should be paid at least the minimum wage in the country of the data collector. 
    \end{itemize}

\item {\bf Institutional review board (IRB) approvals or equivalent for research with human subjects}
    \item[] Question: Does the paper describe potential risks incurred by study participants, whether such risks were disclosed to the subjects, and whether Institutional Review Board (IRB) approvals (or an equivalent approval/review based on the requirements of your country or institution) were obtained?
    \item[] Answer: \answerNA{} 
    \item[] Justification: The paper does not involve any research with human subjects or crowdsourcing activities. All experiments were conducted using publicly available datasets and do not require IRB approval.
    \item[] Guidelines:
    \begin{itemize}
        \item The answer NA means that the paper does not involve crowdsourcing nor research with human subjects.
        \item Depending on the country in which research is conducted, IRB approval (or equivalent) may be required for any human subjects research. If you obtained IRB approval, you should clearly state this in the paper. 
        \item We recognize that the procedures for this may vary significantly between institutions and locations, and we expect authors to adhere to the NeurIPS Code of Ethics and the guidelines for their institution. 
        \item For initial submissions, do not include any information that would break anonymity (if applicable), such as the institution conducting the review.
    \end{itemize}

\item {\bf Declaration of LLM usage}
    \item[] Question: Does the paper describe the usage of LLMs if it is an important, original, or non-standard component of the core methods in this research? Note that if the LLM is used only for writing, editing, or formatting purposes and does not impact the core methodology, scientific rigorousness, or originality of the research, declaration is not required.
    \item[] Answer: \answerNA{} 
    \item[] Justification: The research does not involve the use of large language models (LLMs) as an important, original, or non-standard component of the core methodology. Any language model use was limited to general writing assistance and did not influence the scientific content, experiments, or technical contributions of the paper.
    \item[] Guidelines:
    \begin{itemize}
        \item The answer NA means that the core method development in this research does not involve LLMs as any important, original, or non-standard components.
        \item Please refer to our LLM policy (\url{https://neurips.cc/Conferences/2025/LLM}) for what should or should not be described.
    \end{itemize}

\end{enumerate}

\end{document}